\documentclass[10pt,journal,compsoc]{IEEEtran}

\ifCLASSOPTIONcompsoc
  \usepackage[nocompress]{cite}
\else
  \usepackage{cite}
\fi
\ifCLASSINFOpdf
  \usepackage[pdftex]{graphicx}
  \graphicspath{{./figures}}
  \DeclareGraphicsExtensions{.pdf,.jpeg,.png}
\else
\fi
\usepackage{amsmath}
\usepackage{hyperref}

\usepackage{bm}
\usepackage{amssymb}
\usepackage[ruled, noline, longend]{algorithm2e}

\usepackage{array}

\ifCLASSOPTIONcompsoc
  \usepackage[caption=false,font=footnotesize,labelfont=sf,textfont=sf]{subfig}
\else
  \usepackage[caption=false,font=footnotesize]{subfig}
\fi
\usepackage{url}

\usepackage{tikz}
\usepackage{collcell}

\usepackage{multirow}
\usepackage{booktabs}
\usepackage{color}
\usepackage{pifont}
\usepackage{changepage}

\newcommand*{\MinNumber}{0.0}%
\newcommand*{\MidNumber}{10.0} %
\newcommand*{\MaxNumber}{50.0}%
\newcommand*{\LowNumber}{5.0}%

\definecolor{Gray}{gray}{0.8}

\newcommand{\ApplyGradient}[1]{%
        \ifdim #1 pt > \MidNumber pt
            \pgfmathsetmacro{\PercentColor}{max(min(100.0*(#1 - \MidNumber)/(\MaxNumber-\MidNumber),100.0),0.00)} %
            \textcolor{red!\PercentColor!orange}{#1}
        \fi

        \ifdim #1 pt < \MidNumber pt

            \ifdim #1 pt > \LowNumber pt

                  \pgfmathsetmacro{\PercentColor}{max(min(100.0*(\MidNumber - #1)/(\MidNumber-\MinNumber),100.0),0.00)} %
                \textcolor{green!\PercentColor!orange}{\textbf{#1}}

                \else

            \pgfmathsetmacro{\PercentColor}{max(min(100.0*(#1 - \LowNumber)/(\MinNumber-\LowNumber),100.0),0.00)} %
            \textcolor{blue!\PercentColor!green}{\textbf{#1}}

        \fi

        \fi
        }

        \newcommand{\Black}[1]{%
                    \textcolor{black}{#1}

                }

\newcolumntype{R}{>{\collectcell\Black}{r}<{\endcollectcell}}

\begin{document}

\title {A Comprehensive Evaluation on Multi-channel Biometric Face Presentation Attack Detection}

\author{Anjith George,
        David Geissb\"uhler,
        and~S\'ebastien~Marcel,~\IEEEmembership{Senior~Member,~IEEE}
\IEEEcompsocitemizethanks{
\IEEEcompsocthanksitem
 Anjith George, David Geissb\"uhler and~S\'ebastien Marcel are with the
  Idiap Research Institute, Switzerland, e-mails: \{anjith.george, david.geissbuhler, sebastien.marcel\}@idiap.ch}
}

%
%

\markboth{Journal of \LaTeX\ Class Files,~Vol.~14, No.~8, August~2015}%
{George \MakeLowercase{\textit{et al.}}:A Comprehensive Evaluation on Multi-channel Biometric Face Presentation Attack Detection}
%



\IEEEtitleabstractindextext{%

\begin{abstract}
  The vulnerability against presentation attacks is a crucial problem undermining the wide-deployment of face recognition systems. Though presentation attack detection (PAD) systems try to address this problem, the lack of generalization and robustness continues to be a major concern. Several works have shown that using multi-channel PAD systems could alleviate this vulnerability and result in more robust systems. However, there is a wide selection of channels available for a PAD system such as RGB, Near Infrared, Shortwave Infrared, Depth, and Thermal sensors. Having a lot of sensors increases the cost of the system, and therefore an understanding of the performance of different sensors against a wide variety of attacks is necessary while selecting the modalities. In this work, we perform a comprehensive study to understand the effectiveness of various imaging modalities for PAD.  The studies are performed on a multi-channel PAD dataset, collected with 14 different sensing modalities considering a wide range of 2D, 3D, and partial attacks. We used the multi-channel convolutional network-based architecture, which uses pixel-wise binary supervision.  The model has been evaluated with different combinations of channels, and different image qualities on a variety of challenging known and unknown attack protocols. The results reveal interesting trends and can act as pointers for sensor selection for safety-critical presentation attack detection systems. The source codes and protocols to reproduce the results are made available publicly making it possible to extend this work to other architectures.

\end{abstract}

}

\maketitle

\IEEEdisplaynontitleabstractindextext

%
\IEEEpeerreviewmaketitle

\IEEEraisesectionheading{\section{Introduction}\label{sec:introduction}}
Biometric systems enable a secure and convenient way for access control. Among the biometric systems, face recognition systems has received significant attention in the past years due to its ease of use and non-contact nature. Even though face recognition systems have achieved human parity in ``in the wild'' datasets \cite{learned2016labeled}, their vulnerability to presentation attacks limits its reliable use in secure applications \cite{marcel2014handbook,handbook2}. The ISO standard \cite{ISO1} defines a presentation attack as ``a presentation to the biometric data capture subsystem with the goal of interfering with the operation of the biometric system''. For example, the presentation of a photo of a person in front of a face recognition system constitutes a 2D presentation attack. The attacks could be simple 2D prints or replays, or more sophisticated 3D masks and partial masks. There are many works in the literature that shows the vulnerability of face recognition systems against presentation attacks \cite{kose-icassp-2013} \cite{hadid-cvprw-2014} \cite{chingovska-frais-2016} \cite{mohammadi-iet-2017} \cite{bhattacharjee-btas-2018}.

\begin{figure}[ht!]
\centering
\includegraphics[width=0.95\linewidth]{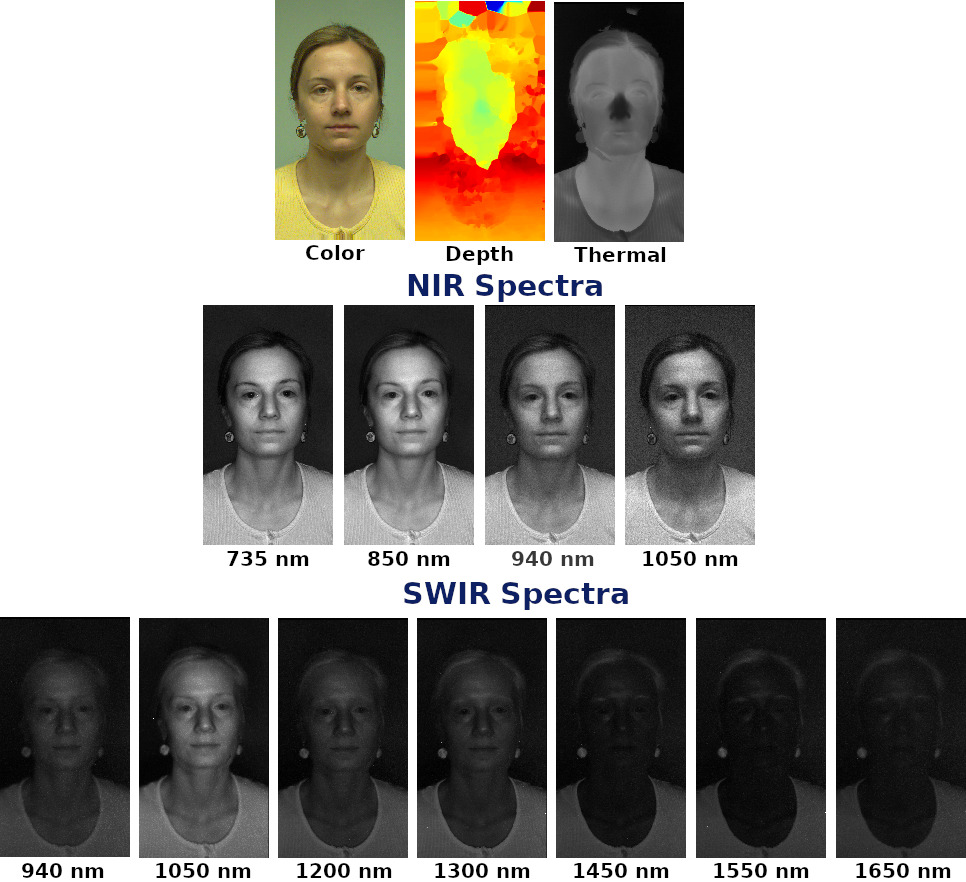} 
\caption{Different channels from HQ-WMCA dataset, first row shows the color, depth and thermal channels; second row shows the channel-wise normalized images from four \textit{NIR} wavelengths, and the third row shows the channel-wise normalized images from different \textit{SWIR} wavelengths. Note that the wavelengths 940nm and 1050nm were captured by both NIR and SWIR sensors.}
\label{fig:channels}
\end{figure}

To alleviate this vulnerability, there have been several works on presentation attack detection, which tries to protect biometric systems against spoofing attempts  \cite{galbally-access-2014}, \cite{li-iet-2018,yeh2018face}. The majority of the literature in face PAD focuses on data collected with RGB channel \cite{liu2018learning,yang2019face,george2019deep,yu2020searching,george2021effectiveness,sun2020face}. Though this is important to have methods that work with legacy RGB data, it is clear that the accuracy of these PAD methods against even print or replay attacks has limits, in addition to poor generalization capacity \cite{de2013can,purnapatra2021face}. Several approaches such as multi-adversarial deep domain generalization \cite{shao2019multi,jia2020single} and disentangled representation learning \cite{wang2020cross,wang2020unsupervised}, has been proposed to improve the generalization of face PAD methods. However, detecting attacks becomes more challenging as the quality of presentation attacks improves over time \cite{9218954,liu2019deep,george2021effectiveness}.

Methods based on extended range imaging offers an alternative approach to improve the robustness of PAD systems. It has been shown that mere fusion of multi-channel systems could readily improve the results as compared to \textit{RGB} only counterparts \cite{george_mccnn_tifs2019,wang2019multi,george2020face,heusch2020deep,nikisins2019domain,parkin2019recognizing,george_mcocm_tifs2020,zhang2019dataset,wang2019multi}. Indeed, it can be seen that its harder to attack systems relying on multiple channels of information. Recently, there have been several CNN based methods that leverage multi-channel information for PAD. In addition to legacy RGB data, several works have suggested the use of depth, thermal \cite{agarwal2017face,george_mccnn_tifs2019, bhattacharjee2017you,george2021multi,george2021cross}, near-infrared (\textit{NIR}), shortwave infrared (\textit{SWIR}) \cite{heusch2020deep, steiner2016design}, laser speckle imaging, light field imaging \cite{raghavendra2015presentation, raghavendra2017vulnerability} and so on. However, the effectiveness of these channels on a wide variety of attacks has not been studied comprehensively. The channels used should contain discriminative information useful for PAD, and at the same time reduce redundant information. Multi-channel PAD methods provide better performance and robustness in safety-critical applications. However, for practical deployment scenarios adding more channels increases the cost of the systems. An understanding of the performance of various channels is essential to decide which channels need to be used for a robust PAD system. Indeed, the channels used should perform well in both known and unseen attack scenarios, while remaining cost-effective.

In this work, the objective is to analyze the performance of multi-channel presentation attack detection systems against a wide variety of attacks. Specifically, we consider the static scenario where PAD output can be obtained with a single sample, consisting of either single or multi-channel images. An extensive set of experiments have been performed to understand the change in performance and robustness of the PAD systems with different combinations of channels and image resolutions. We use our recently publicly available dataset \textit{HQ-WMCA} \cite{heusch2020deep} for this analysis, which contains a wide range of attacks collected with a wide variety of acquisition devices. In fact, for each sample, data collected from 14 distinct imaging modalities are available with the dataset. We hope that the results and insights from the analysis could act as pointers in selecting the channels to be used with robust PAD systems.

The objectives of this work are summarized as follows:

\begin{itemize}
	\item Identify important subsets of channels, which contain discriminative information for PAD (as cost vs performance trade-off) by analyzing the performance in a wide range of experimental protocols.
	\item Identify the minimum quality requirements, in terms of resolution for the channels used.
	\item Evaluate the robustness of various channels against unseen attacks.
\end{itemize}

Here, the term \textit{channel} refers to various imaging modalities like RGB, Thermal, Depth, NIR, and SWIR spectra.

For this work, we use the multi-channel extension of Pixel-wise Binary Supervision \cite{george2019deep} as presented in our recent work \cite{heusch2020deep} adapting it to all channels available in \textit{HQ-WMCA} dataset. We selected this PAD model due to its capacity to accept a varying number of input channels, ease of implementation, and excellent performance against a wide range of attacks. In our previous work \cite{heusch2020deep}, we mainly analyzed various combinations of
SWIR channels and their effectiveness in various known attack protocols. In the current contribution, we comprehensively analyze the performance of multiple channels, different channel combinations, different sensor resolutions, different fusion strategies, and different CNN-based as well as feature-based methods. Further, in addition to studies in known attack protocols, we also present experiments in challenging leave-one-
out protocols, enabling us to evaluate the performance of the
algorithms in unseen attack scenarios. In essence, we comprehensively compare the effectiveness of different sensing modalities,  considering a cost versus performance trade-off.

Additionally, the source code to reproduce the results are available publicly \footnote{ Available upon acceptance } to extend the present work with any other architecture. The framework and methodology used for this study are detailed in Section \ref{sec:experiments}.

The rest of the paper is organized as follows. Section \ref{sec:scalability} describes the different components of the study performed. Section \ref{sec:database} presents the details of the hardware used, the dataset, and the newly introduced protocols. Section \ref{sec:pad} describes the details of the PAD framework used.  Section \ref{sec:experiments} consists of all the experiments and results. Finally, Section \ref{sec:conclusion} concludes the paper.

%
%

\section{Details of Study}
\label{sec:scalability}
The objective of this study is to understand the performance of a PAD system with various combinations of imaging modalities and with different quality in terms of image resolutions. All the combinations are evaluated in the created protocols to understand their performance in known and unseen attack performance. The motivations of this study are described in the following section.

From a practical viewpoint, it is challenging to collect data from a lot of sensors at once as it increases the cost significantly. For instance, the cost of Xenics Bobcat SWIR sensor is around \$31500 and the cost of Xenics Gobi thermal camera is around \$10500. There could be correlated factors among the different sensors and channels. There could be channels with no discriminative information for PAD, and it might result in over-fitting. Getting rid of redundant information might make a PAD model less prone to over-fitting and more robust. A larger number of channels increase the parameters of the model, which again increases the chance of over-fitting. This is especially true for cases with a limited amount of training data. In addition, the image resolution required for the PAD system to achieve a prescribed performance may be lower than what is used.

In summary, while designing a sensor suite for real-world deployment, the system has to meet certain cost-vs-performance criteria. However, to the best of the knowledge of the authors, there exists no previous work which studies the effectiveness of various combinations of channels against a challenging variety of attacks in a comprehensive manner.  This warrants a rigorous analysis of the performance of the PAD system with these covariates.

Specifically, the important questions which this study is trying to reveal are:

\begin{itemize}
\item Which channels or combinations of channels contain discriminative information for PAD?
\item How does the performance of the PAD system change with the change in the resolution of images?
\item Which channels or combinations of channels are robust against unseen attacks?
\item Is the fusion of channels at the model-level better than fusion at the score level?
\end{itemize}

\begin{figure*}[ht!]
\centering
\includegraphics[width=0.99\textwidth]{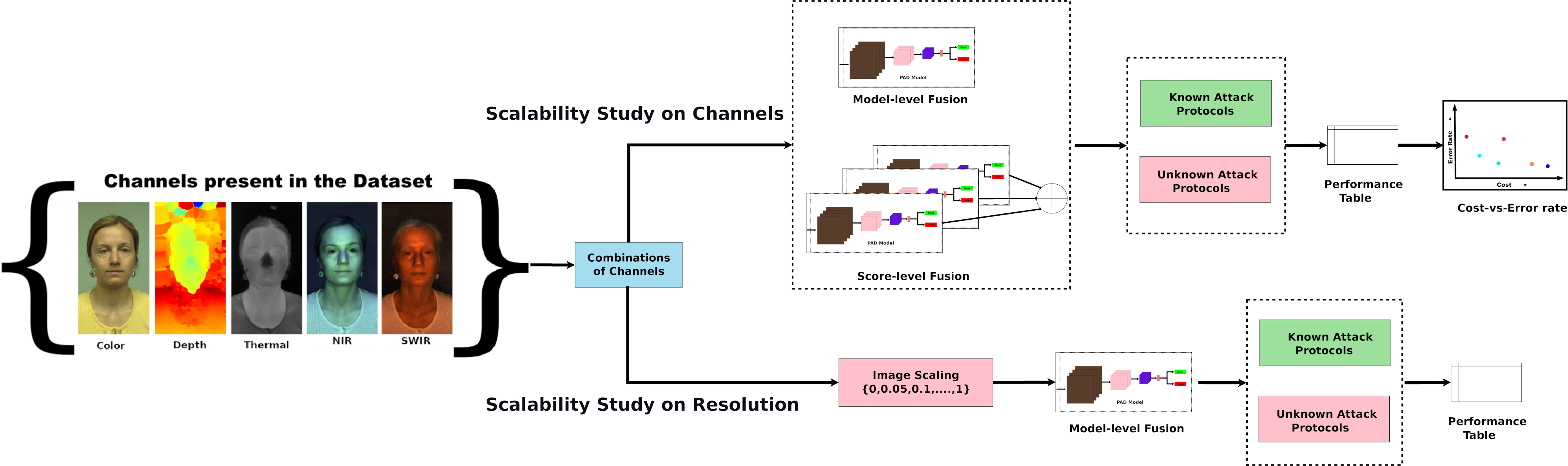}
\caption{The framework for the channel selection and image resolution selection study. Each set of experiments is repeated in several challenging protocols to evaluate the robustness.}
\label{fig:scalability_diagram}
\end{figure*}

To answer these questions, we performed several sets of experiments, and the details of each analysis are provided below. A flowchart of the study is provided in Fig. \ref{fig:scalability_diagram}.

\subsection{Channel Selection studies}
\label{sec:channel-selection}

The objective of this study is to identify important channels for deploying a reliable PAD system. The cost of using all the channels in a PAD system is high, and hence the understanding of different combinations of these channels as a cost vs performance trade-off could be useful in selecting sensors for practical deployment scenarios.

The PAD model used in this study can be adapted to a different number of input channels. In the channel selection study, we perform experiments with a different combination of channels as the input to the model.

As described in the preprocessing sub-section \ref{subsec:preprocess}, the input after the preprocessing stage is a data cube of size $224 \times 224 \times 16 $ which consists of all channels present in the dataset, i.e., \textit{RGB}-Depth-Thermal-\textit{NIR}(4)-\textit{SWIR}(7) stacked together depth-wise. Now, if we want to perform the experiment with \textit{RGB} alone, then we keep only the first 3 channels and run the complete experimental pipeline (training the PAD model, evaluation, and scoring). Similarly, if we want to perform the experiment with \textit{RGB} and \textit{SWIR} (\textit{RGB-SWIR}), we stack the first 3 channels and the last 7 channels and run the experimental pipeline. In a similar manner, we can perform experiments with different combinations of channels present in the dataset. The results from this experiment could point to the effectiveness of a PAD model with a particular channel or a combination of channels for different protocols, emulating practical scenarios.

\subsection{Score Fusion Experiments}

In the Channel Selection studies (Subsection \ref{sec:channel-selection}), when two channels are used together, they are stacked together at the input (fusion at the model level). Another way to combine different models is to perform score level fusion \cite{chingovska2013anti}. If we want to perform a fusion of \textit{RGB-SWIR}, we train two different PAD models for \textit{RGB} and \textit{SWIR} separately, and score level fusion is performed on the scores returned by each of the models. In addition to score fusion ,we perform experiments with feature fusion as well. The objective of these set of experiments is to analyze the performance of score level fusion and feature level fusion, as opposed to early fusion as used in the channel selection study.

\subsection{Effect of changing quality (image resolution)}

From a practical deployment point of view, the cost of the sensors varies significantly with image resolution. This is particularly evident when considering the SWIR channel, as high-resolution SWIR sensors are rather expensive. Hence we perform a set of experiments to evaluate the change in performance with respect to image resolution. To emulate a lower resolution sensor, we have down-sampled the original image by various scaling factors. We repeated all the experiments with the scaled images, so as to emulate the PAD performance with a lower resolution sensor.

\subsection{Unseen attack robustness}

One important requirement of the PAD system is the robustness against unseen attacks. This is particularly important since at the time of training a PAD system the designer may not know of all the types of novel attacks the system could encounter in a real-world scenario. To understand the robustness of the system to such adverse conditions, we emulate the performance of the system on previously unseen attacks by systematically removing some attacks in the training set. These systems are then exposed to attacks that were not seen in the training phase. This analysis reveals important aspects of different channels in terms of unseen attack robustness.


\section{The HQ-WMCA Dataset}
\label{sec:database}

\begin{figure*}[ht!]
\centering
  \subfloat[]{\includegraphics[height=2.3cm]{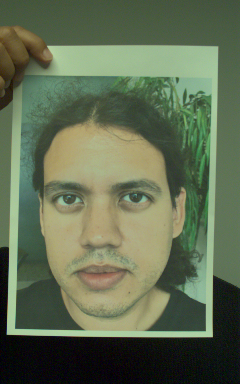}}%
\hfil
  \subfloat[]{\includegraphics[height=2.3cm]{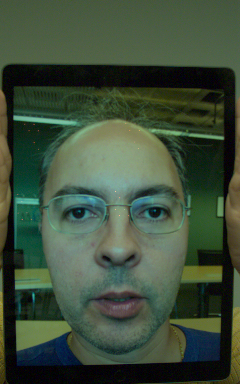}}%
\hfil
  \subfloat[]{\includegraphics[height=2.3cm]{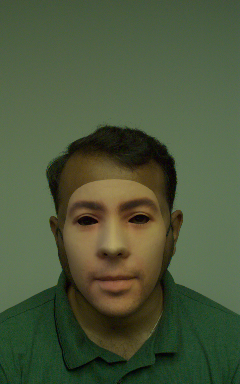}}%
\hfil
  \subfloat[]{\includegraphics[height=2.3cm]{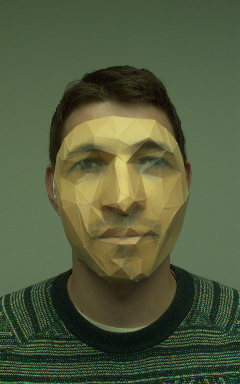}}%
\hfil
  \subfloat[]{\includegraphics[height=2.3cm]{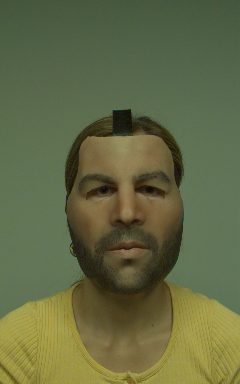}}%
\hfil
  \subfloat[]{\includegraphics[height=2.3cm]{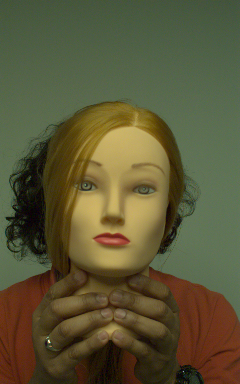}}%
\hfil
  \subfloat[]{\includegraphics[height=2.3cm]{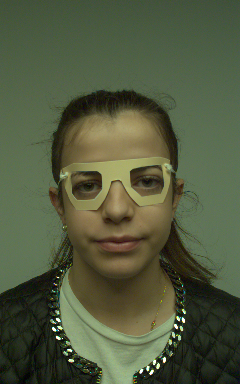}}%
\hfil
  \subfloat[]{\includegraphics[height=2.3cm]{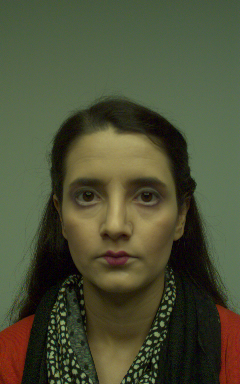}}%
\hfil
  \subfloat[]{\includegraphics[height=2.3cm]{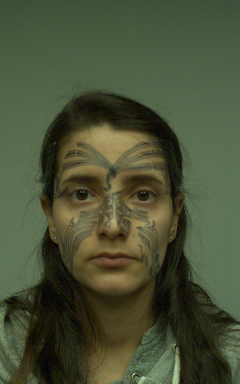}}%
\hfil
  \subfloat[]{\includegraphics[height=2.3cm]{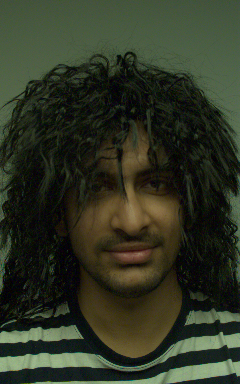}}%
  \caption{(a) Print, (b) Replay, (c) Rigid mask, (d) Paper mask, (e) Flexible mask, (f) Mannequin,
          (g) Glasses, (h) Makeup, (i) Tattoo and (j) Wig. Image taken from \cite{heusch2020deep}.}
\label{fig:attacks}
\end{figure*}

The dataset used in our study is the recently available High-Quality Wide Multi-Channel Attack dataset (HQ-WMCA) \cite{heusch2020deep, Mostaani_Idiap-RR-22-2020}. The dataset is publicly available \footnote{\url{https://www.idiap.ch/dataset/hq-wmca}}.  The dataset contains a wide variety of 2D, 3D, and partial attacks. Some samples of the attacks are shown in Fig. \ref{fig:attacks}. The attacks consist of both obfuscation and impersonation attacks. For each video, the data collected includes \textit{RGB}, \textit{NIR} (4 wavelengths), \textit{SWIR} (7 wavelengths), thermal, and depth channels.

\begin{figure}[h]
\centering
\includegraphics[width=0.3\textwidth]{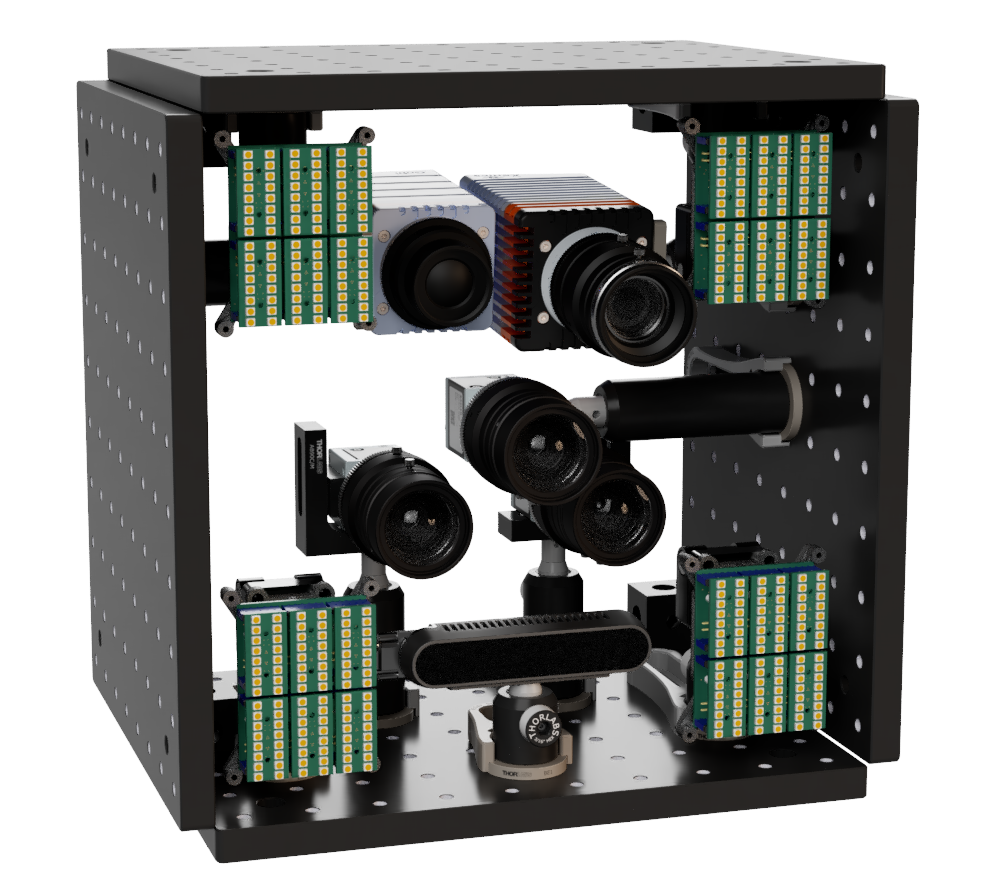}
\caption{Sensor suite used for the data collection in HQ-WMCA dataset.}
\label{fig:face-station}
\end{figure}

The sensor suite used for the data collection is shown in  Fig. \ref{fig:face-station}. 
The details and the specifications of the sensors are shown in  Table \ref{tab:sensors}.

\begin{table}[h]
\centering
\caption{Specifications of the sensors and the approximate cost.}
\label{tab:sensors}
\resizebox{0.49\textwidth}{!}{
\begin{tabular}{llccc}
\toprule
Sensor name            & Modality & Resolution       & Frame rate & Cost (USD) \\
\midrule
Basler acA1921-150uc   & Color   & 1920$\times$1200  & 30         & 1,300  \\
Basler acA1920-150um   & NIR     & 1920$\times$1200  & 90         & 1,250 \\
Xenics Bobcat-640-GigE & SWIR    & 640$\times$512    & 90         & 31,500 \\
Xenics Gobi-640-GigE   & Thermal & 640$\times$480    & 30         & 10,500 \\
Intel RealSense D415   & Depth   & 720$\times$1280   & 30         & 150\\
\bottomrule
\end{tabular}
}
\end{table}

The data was collected at different sessions, and under different lighting conditions. The details of the data collection can be found in our previous work \cite{heusch2020deep}.

The high quality and the wide variety of channels, along with the variety of attacks present in the \textit{HQ-WMCA} dataset make it a suitable candidate for performing the evaluation. The normalized images from different channels are shown in Fig. \ref{fig:channels}.

\subsection{Reprojection-based registration}
\label{subsec:reprojection}
As discussed earlier, there are several sensors present in the sensor suite which captures information in a time-synchronized manner. However, to utilize the information from all channels together, the images captured by the sensors should be registered accurately. A 3D Re-projection method is used for the robust registration between the sensors.

Re-projection is a process allowing synchronized video streams from different cameras or sensors to be precisely aligned despite cameras being at different spatial positions.
For this algorithm to be functional, the precise relative rotations and positions of the cameras, also known as extrinsic parameters, as well as the lens distortion coefficients and projection matrix, or intrinsic parameters, have to be inferred. This is achieved by capturing a series of images of a chosen pattern, for instance, a flat checkerboard target, by all cameras. Both intrinsic and extrinsic parameters can be calculated by detecting positions of the target's features on the images, and using OpenCV's camera calibration module \cite{opencv_library}.

\begin{figure}[h]
\centering
\includegraphics[width=0.95\linewidth]{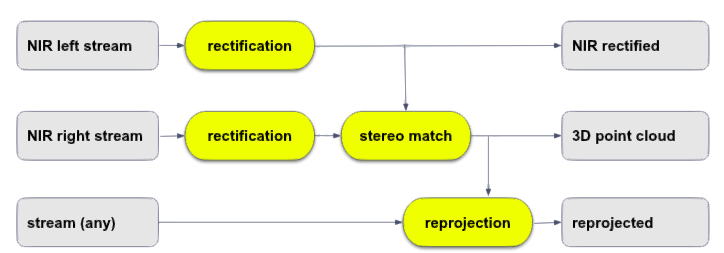} 
\caption{Diagram of the reprojection based registration of images from multiple sensors.}
\label{fig:reproj_flowchart}
\end{figure}

\begin{figure}[h]
\centering
\includegraphics[width=0.95\linewidth]{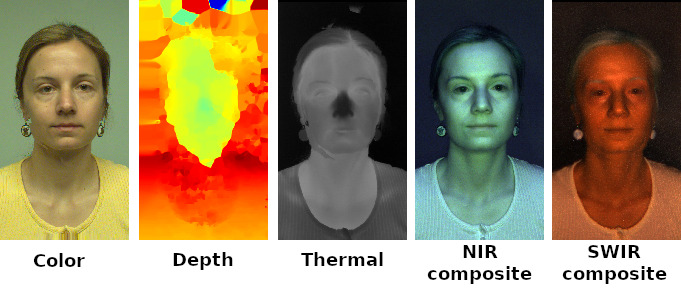}
\caption{Sample results from reprojection based alignment color channel, stereo depth, thermal channel, NIR composite, and SWIR composite. }
\label{fig:reproj_channels}
\end{figure} 

\begin{figure}[t!]
\centering
\includegraphics[width=0.99\linewidth]{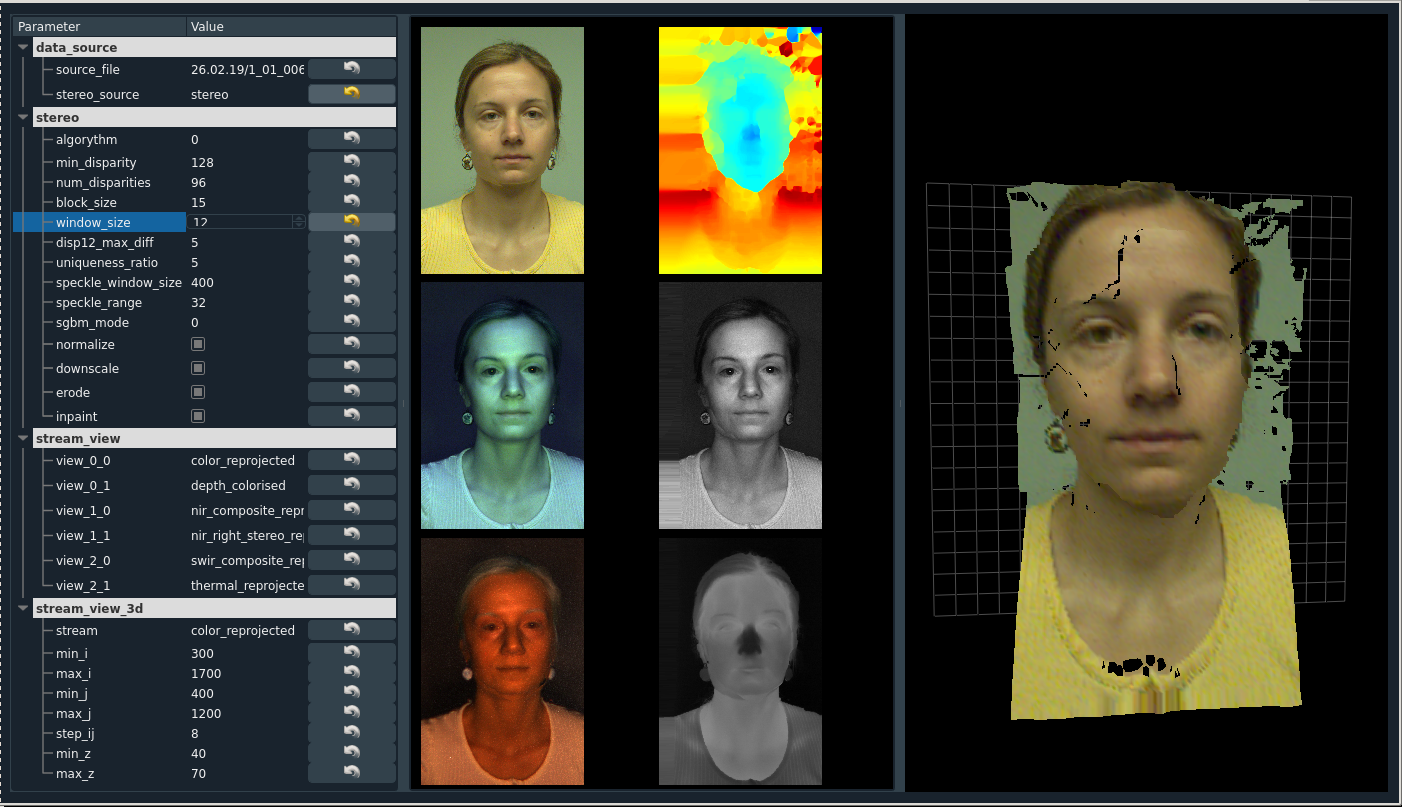}
\caption{Interface developed for visualizing the alignment procedure. The best calibration and parameter tuning can be selected based on this tool. On the right side of the interface, any channel (here RGB) can be projected in 3D and then after rectification reprojected to get same alignment between all channels.}
\label{fig:3dtool}
\end{figure} 

The set of cameras is assumed to have a depth-sensing device, for the present discussion this is achieved by a pair of NIR sensing cameras. The diagram of the registration process is shown in Fig. \ref{fig:reproj_flowchart}. An example of the stereo image computed and registered channels are shown in Fig. \ref{fig:reproj_channels}. The NIR cameras have an extra calibration step, adapting their intrinsic and extrinsic parameters such that they are properly aligned for depth reconstruction, these views are called rectified left and right in the following, respectively. The depth reconstruction is performed using a block matching algorithm \cite{opencv_library}, that measures the disparity between the pair of images. The latter is transformed to a point cloud, yielding a point in 3D space for each pixel in the left image. The re-projection algorithm then proceeds by projecting the point cloud, with points tagged by their coordinates on the left rectified image, on the virtual image plane of the remaining cameras. This procedure thus creates a set of maps from the left rectified image plane to the other cameras' images planes, by inverting these maps the video streams can be precisely aligned together, for instance to the reference left stream. A screen-shot of the graphical user interface (GUI) developed for viewing the alignment process is shown in Fig. \ref{fig:3dtool}.

\subsection{Protocols}

The \textit{HQ-WMCA} dataset is distributed with three different protocols, namely \textit{grand-test}, \textit{impersonation} and \textit{obfuscation}. In each of these protocols, there are three folds, namely train, validation, and test sets. These folds contain both bonafide samples and attacks (Fig. \ref{fig:attacks}) with a disjoint set of identities across the three sets. The bonafide and attacks are roughly equally distributed across the folds and each video contain 10 frames, sampled uniformly from the captured data.

We also noticed that some attacks are questionable and may be considered as ``occluded'' bonafide which could potentially confound the analysis. Specifically, these borderline cases are: \\

\begin{itemize}
  \item \textbf{Wigs}: The wigs do not occlude a lot of face regions in most of the cases, and since the PAD and face recognition module removes the non-face region in the preprocessing stage itself, its effect in spoofing face recognition system is not clear. Hence we removed wigs from all the protocols to avoid any bias.
  \item \textbf{Retro-glasses}: These are similar to normal medical glasses and is identical to a bonafide subject wearing a medical glass and hence we removed this attack from all our new protocols.
  \item \textbf{Light makeup}: In the original data collected, for each subject, for each makeup session, three samples were collected at different levels of makeup, namely level 0, level 1, and level 2 depending on the level of makeup applied to the subject. The level of makeup in level 0 is very insignificant and could be identical to routine makeup present in bonafide samples. Hence we have removed the level 0 makeup samples from the newly created protocols to have a consistent set of ground truths.
\end{itemize}

Three experiment protocols were newly created, as curated versions of the protocols provided with the dataset \cite{heusch2020deep}. The newly created curated protocols are hence appended with ``-c'' to emphasize the difference from the original protocols shipped with the dataset, i.e.,  the adapted and newly created protocols are referred to as, \textit{Grandtest-c}, \textit{Impersonation-c} and \textit{Obfuscation-c} protocols.

In addition, we have also created a set of leave-one-out protocols (\textit{LOO}), to emulate the unseen attack scenarios.

To summarize, the protocols used in the present work are (Table \ref{tab:attacks-distribution}):
\begin{itemize}
    \item \textbf{Grandtest-c}: This is the same grand-test protocol shipped with the \textit{HQ-WMCA} dataset, after removing the borderline cases. Here all the attacks, appear in different folds of the dataset \textit{train}, \textit{validation}, and \textit{test}.

    \item \textbf{Impersonation-c}: This is the same impersonation protocol shipped with the \textit{HQ-WMCA} dataset, after removing the borderline cases. Only attacks for impersonation are present in this protocol, i.e., attacks for which the attacker is trying to authenticate himself as another user. The attacks present in this protocol are shown in Table \ref{tab:attacks-distribution}.

    \item \textbf{Obfuscation-c}: In a similar manner, this is the same obfuscation protocol shipped with the \textit{HQ-WMCA} dataset, after removing the borderline cases. This protocol consists of obfuscation attacks, which refers to attacks in which the appearance of the attacker is altered so as to prevent being identified by a face recognition system. The attacks present in this protocol are makeups, tattoos, and glasses.

    \item \textbf{Leave-One-Out protocol}: A set of different sub-protocols are newly created to emulate unseen attacks. The sub-protocols are created by leaving out each attack category in the training set. There are eight sub-protocols in the \textit{LOO} protocol, which emulate unseen attack scenarios. We start with the splits used in the \textit{grandtest-c} protocol and systematically remove one attack in the \textit{train} and \textit{validation} sets, the \textit{test} set consists of \textit{bonafide} and the attack which was left out. We created \textit{LOO} protocols for the different attacks present in the dataset. And the protocols are listed below:\textit{LOO\_Flexiblemask}, \textit{LOO\_Glasses}, \textit{LOO\_Makeup}, \textit{LOO\_Mannequin}, \textit{LOO\_Papermask}, \textit{LOO\_Rigidmask}, \textit{LOO\_Tattoo} and \textit{LOO\_Replay}. 
    The distribution of attacks in this protocol can be found in Table \ref{tab:bf_pa-distribution}.
\end{itemize}

Overall there are $360,080$ images considering all the modalities after excluding the borderline attacks.


The statistics of bonafide and attacks in each of the folds in each protocol are given in Table \ref{tab:bf_pa-distribution}.

\begin{table*}[h]

  \centering
  \caption{Distribution of bonafide and attack in the various protocols.}

\begin{tabular}{@{}lcccccc@{}}
\toprule
\multirow{2}{*}{\textbf{Protocol}} & \multicolumn{2}{c}{\textbf{Train}} & \multicolumn{2}{c}{\textbf{Validation}} & \multicolumn{2}{c}{\textbf{Test}} \\ \cmidrule(l){2-7}
                          & Bonafide     & Attacks    & Bonafide    & Attacks   & Bonafide    & Attacks    \\ \midrule
Grandtest-c             & 228          & 618        & 145         & 767       & 182         & 632        \\
Impersonation-c           & 228          & 384        & 145         & 464       & 182         & 440        \\
Obfuscation-c             & 228          & 234        & 145         & 303       & 182         & 192        \\ \cmidrule{1-7}
LOO\_Flexiblemask         & 228          & 528        & 145         & 681       & 182         & 48         \\
LOO\_Glasses              & 228          & 582        & 145         & 729       & 182         & 36         \\
LOO\_Makeup               & 228          & 444        & 145         & 526       & 182         & 132        \\
LOO\_Mannequin            & 228          & 598        & 145         & 729       & 182         & 77         \\
LOO\_Papermask            & 228          & 590        & 145         & 743       & 182         & 49         \\
LOO\_Rigidmask            & 228          & 456        & 145         & 649       & 182         & 140        \\
LOO\_Tattoo               & 228          & 594        & 145         & 743       & 182         & 24         \\
LOO\_Replay               & 228          & 582        & 145         & 667       & 182         & 126        \\ \bottomrule
\end{tabular}

\label{tab:bf_pa-distribution}

\end{table*}

The distribution of attack types, in the three main protocols, are shown in Table \ref{tab:attacks-distribution}.

\begin{table}
  \centering
  \caption{Distribution of attacks in the different protocols.}
  \begin{tabular}{llccc}
    \toprule

     &\textbf{Attack type} & \textbf{Train} & \textbf{Validation} & \textbf{Test}\\
  \midrule
\multirow{10}{*}{\textbf{Grandtest-c}}
& Flexiblemask &   90 &   86 &   48 \\
& Glasses      &   36 &   38 &   36 \\
& Makeup       &  174 &  241 &  132 \\
& Mannequin    &   20 &   38 &   77 \\
& Papermask    &   28 &   24 &   49 \\
& Print        &   48 &   98 &    0 \\
& Replay       &   36 &  100 &  126 \\
& Rigidmask    &  162 &  118 &  140 \\
& Tattoo       &   24 &   24 &   24 \\ \cmidrule{2-5}
& \textit{Bonafide}     &  228 &  145 &  182 \\
  \midrule
\multirow{7}{*}{\textbf{Impersonation-c}}
& Flexiblemask &   90 &   86 &   48 \\
& Mannequin    &   20 &   38 &   77 \\
& Papermask    &   28 &   24 &   49 \\
& Print        &   48 &   98 &    0 \\
& Replay       &   36 &  100 &  126 \\
& Rigidmask    &  162 &  118 &  140 \\ \cmidrule{2-5}
& \textit{Bonafide}     &  228 &  145 &  182 \\
  \midrule
\multirow{4}{*}{\textbf{Obfuscation-c}}
& Glasses  &     36 &   38 &    36 \\
& Makeup   &    174 &  241 &   132 \\
& Tattoo   &     24 &   24 &    24 \\ \cmidrule{2-5}
& \textit{Bonafide} &    228 &  145 &   182 \\
\bottomrule
  \end{tabular}
  \label{tab:attacks-distribution}
\end{table}


\section{Presentation Attack Detection Approach}
\label{sec:pad}
\begin{figure*}[ht!]
\centering
\includegraphics[width=0.85\textwidth]{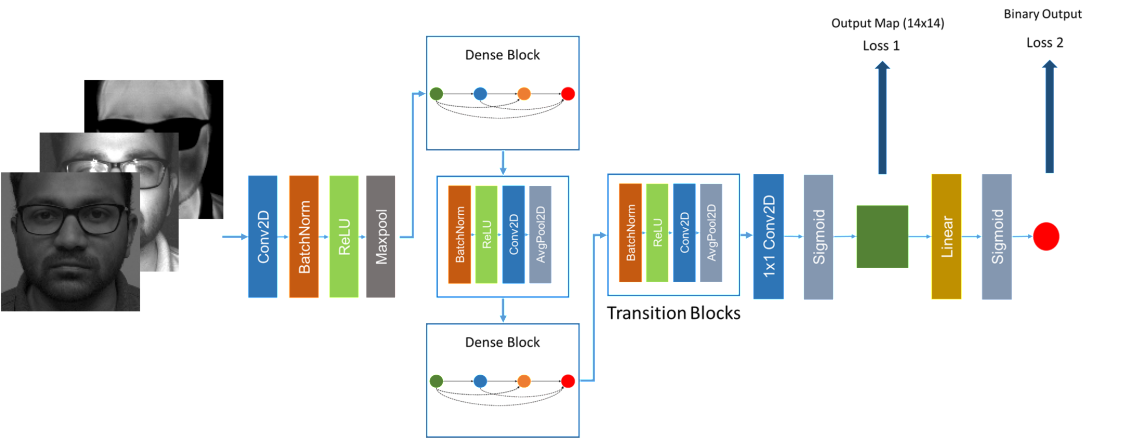}
  \caption{MC-PixBiS architecture with pixel-wise supervision. All the channels are stacked together to form a 3D volume, and fed as the input to the network.
    At train time, both binary and pixel-wise binary loss functions are used to supervise learning. At test time, the average of the score map is used as the final PAD score.}

\label{fig:pixbis}
\end{figure*}

The objective of this work is to analyze the performance of a PAD system against a wide variety of attacks, with different combinations of channels.  This analysis requires a competitive framework for performing the experiments. CNN based methods generally outperform feature-based baselines in the PAD task and it is more evident in multi-channel scenarios as well \cite{george_mccnn_tifs2019, george2020face}. Hence, we perform the experiments on a competitive CNN architecture. We have selected \textit{MC-PixBiS} \cite{heusch2020deep}, due to its excellent PAD performance and ease of implementation.  The training and inference stages are rather straightforward, and the framework itself is adaptable to a varying number of input channels, making the experiments with different combinations of channels possible. Moreover, the source code for the architecture is available publicly. The details of the architecture are outlined in the following sections.

\subsection{Network Architecture}
\label{subsec:arch}

The architecture used is the multi-channel extension of \cite{george2019deep}, as in \cite{heusch2020deep}. From our previous work, it was observed that out of the architectures we considered, \textit{MC-PixBiS} obtained much better accuracy, with a relatively smaller model size. Hence, the rest of the analysis in this work utilizes the \textit{MC-PixBiS} architecture proposed in our previous work \cite{heusch2020deep}.

The main idea in \textit{MC-PixBiS} is to use pixel-wise supervision as auxiliary supervision for training a multi-channel PAD model. The pixel-wise supervision forces the network to learn shared representations, and it acts like a patch wise method (see Figure~\ref{fig:pixbis}) depending on the receptive field of the convolutions and pooling operations used in the network.

The method proposed in \cite{wang-eccv-2016} is used to initialize the newly formed first layers, i.e. averaging the filters in the first layer and replicating the weights for different channels. This makes the network very flexible to perform the experiments since we can arbitrarily change the number of input channels and the number of filters in the first convolutional layer. The number of parameters of the rest of the layers remains identical for all the experiments. An increase in the input channel changes the size of the kernels of the first convolutional layer, and hence the parameter size does not increase drastically with newly added channels.

The details of the architecture except for the first layer can be found in \cite{george2019deep}. The output from the eighth layer is a map of size $14\times14$ with 384 features. A $1 \times 1$ convolution layer is added along with sigmoid activation to produce the binary feature map and another fully connected layer on top for binary supervision.

The loss to be minimized in the training is a weighted combination of the binary and pixel-wise binary loss:

\begin{equation}
\label{eq:pixbis-loss}
\mathcal{L} = 0.5 \mathcal{L}_{pix}+ 0.5 \mathcal{L}_{bin}
\end{equation}

where $\mathcal{L}_{pix}$ is the binary cross-entropy loss applied to each element of the $14 \times 14$ score map and $\mathcal{L}_{bin}$ is
the binary cross-entropy loss on output after the fully connected layer.

At test time, the average of the score map is used as the final PAD score.

\subsection{Preprocessing}
\label{subsec:preprocess}

The preprocessing stage assumes the data from different channels to be registered to each other. This is made sure by the reprojection method described in section \ref{subsec:reprojection}. After acquiring the registered data, the face images are aligned with face detection and landmark detection. MTCNN \cite{zhang2016joint} face detector was used in \textit{RGB} channel to localize face and facial landmarks. This is followed by a warping based alignment so that the positions of eyes are at the predefined position, post the transformation. The aligned face images are resized to a fixed resolution of  $224 \times 224$. This process is repeated for each channel individually.

In addition to spatial alignment, another normalization is applied to different channels in a case by case basis. The objective of this normalization is to convert all channels to an 8-bit format preserving the maximum dynamic range. The RGB images do not need this normalization stage since it is already in an 8-bit format. The depth and thermal channels are converted to 8-bit range using Median of Absolute Deviation (MAD) based normalization as described in our earlier work \cite{george_mccnn_tifs2019}. However, \textit{NIR} and \textit{SWIR} channels required a special treatment due to their spectral nature. Specifically, we have performed pixel-wise unit normalization for \textit{NIR} and \textit{SWIR} spectra.

Consider the SWIR spectra cube $S$ of size $W \times H \times C$, where $W$ and $H$ are the width and height and $C$ the number of wavelengths ($C=7$). Now a pixel-spectra $\vec{X}=S(i,j,1..C)$, a vector of dimension $C$ ($\vec{X} \in R^C$).

Now each of this pixel $\vec{X}$ is divided by the norm of this vector to form the normalized spectra.
\begin{equation}
\vec{\hat{X}}=\frac{\vec{X}}{\left \| X \right \|_{2}}
\end{equation}

This normalization is performed independently for \textit{NIR} and \textit{SWIR} channels.

We stack all the normalized channels (i.e., \textit{RGB}, depth, thermal, \textit{NIR}, \textit{SWIR}) into a 3D volume of dimension $224 \times 224 \times 16$, which is fed as the input to the network. This is equivalent to the early fusion approach where channels are stacked at the input level.

\subsection{Implementation details}

We used data augmentation using random horizontal flips with a probability of 0.5. Adam Optimizer \cite{kingma2014adam} was used as the optimizer with a learning rate of $1\times10^{-4}$ and a weight decay parameter of  $1\times10^{-5}$. We trained the network for 30 epochs on a GPU with a batch size of 32. The final model was selected based on the best validation loss in each of the experiments. The framework was implemented using PyTorch \cite{paszke2017automatic} and all the experimental pipelines were implemented using Bob toolbox \cite{anjos-icml-2017}. The source code to reproduce the results is available in the following link \footnote{Available upon acceptance.}.


\section{Experiments \& Results}
\label{sec:experiments}
This section provides details about the experiments and the analysis of the results.

\subsection{Metrics}
We have used the standardized ISO metrics for the performance comparison of the various configurations.
The Attack Presentation Classification Error Rate (APCER) which is defined as probability of a
successful attack:
\begin{equation}
  APCER = \frac{\text{ \# of falsely accepted attacks}}{\text{ \# of attacks}}
\end{equation}
and Bonafide Presentation Classification Error Rate (BPCER) is the probability of bonafide sample being classified as an attack.
\begin{equation}
  BPCER = \frac{\text{ \# of rejected real attempts}}{\text{ \# of real attempts}}
\end{equation}

Since we are dealing with a wide range of attacks, and comparison purposes we group all the attacks together while computing the
Average Classification Error Rate (ACER), which is different from  the ISO/IEC 30107-3 standard \cite{ISO-30107-3}, where
each attack needs to be considered separately. However, we report in ACER in an aggregated manner, since its easier to understand
the variation in performance over a large search space of configurations.
\begin{equation}
\label{eq:ACER}
  ACER = \frac{APCER + BPCER}{2} \quad \textrm{[\%]}
\end{equation}

In this work, a BPCER threshold of 1\% is used in the \textit{dev} set to find the threshold. This threshold is applied to the \textit{test} set to
compute the ACER in the \textit{test} set, as in \cite{george_mccnn_tifs2019}.

\subsection{Channel-wise Experiments}
\label{sec:channel_exp}

\begin{table*}[ht!]
\caption{Performance of different models in the \textit{Grandtest-c}, \textit{Impersonation-c} and \textit{Obfuscation-c} protocols in \textit{HQ-WMCA}, with reprojection and unit spectral normalization. ACER in the \textit{test} set corresponding to BPCER 1\% threshold in \textit{validation} set is shown in the table. The notation ``D'' indicates ``D-Stereo'' in all the experiments.}
\centering
\begin{tabular}{lRRRR}
\toprule

Channels &  \multicolumn{1}{c} {Grandtest-c} &  \multicolumn{1}{c} {Impersonation-c} &  \multicolumn{1}{c} {Obfuscation-c} &  \multicolumn{1}{c} {Mean} \\
\midrule
              RGB  &         4.6 &            0.6 &         14.8 &   6.6 \\      
       D-Stereo  &         26.7&            3.8 &         48.7 &  26.4 \\
        D-Intel  &         29.0&            10.2&         41.8 &   27.0\\
        T &        44.9 &            0.9 &         50.0 &  31.9 \\
        NIR  &         9.7 &            \textbf{0.1} &         39.6 &  16.4 \\
        \textbf{SWIR} &         \textbf{4.1} &            1.8 &          \textbf{9.2} &   \textbf{5.0} \\
                  \midrule
 \textbf{RGB-SWIR} &         0.3 &            2.0 &          \textbf{0.0} &   \textbf{0.7} \\
      RGB-D-T-SWIR &\textbf{0.0} &            2.5 &          \textbf{0.0} &   0.8 \\
  RGB-D-T-NIR-SWIR &\textbf{0.0} &            0.7 &          2.3 &   1.0 \\
           RGB-NIR &         0.7 &   \textbf{0.0} &          7.4 &   2.7 \\
          NIR-SWIR &         2.7 &            0.4 &          8.4 &   3.8 \\
             RGB-D &         6.4 &            1.7 &         12.4 &   6.8 \\
       RGB-D-T-NIR &         3.4 &            0.1 &         17.4 &   6.9 \\
             RGB-T &         6.9 &            2.5 &         17.6 &   9.0 \\
           RGB-D-T &         6.2 &            3.4 &         20.1 &   9.9 \\
               D-T &        17.0 &            4.1 &         49.9 &  23.6 \\

\bottomrule
\end{tabular}
\label{tab:ablation_chanels}
\end{table*}

While the \textit{HQ-WMCA} dataset contains a lot of channels, many of them could be redundant for the PAD task. Also, considering the cost of hardware, it is essential to identify the important channels; respecting the cost factor. Given the set of channels, we perform an ablation study in terms of channels used to identify the minimal subset of channels from the original set to achieve acceptable PAD performance.

The channels present in our study are listed below:
\begin{itemize}
    \item \textbf{RGB}: Color camera at full HD resolution
    \item \textbf{Thermal}: VGA resolution
    \item \textbf{Depth} from Intel Realsense/ \textbf{Stereo} from \textit{NIR} Left/Right Pair. For this study, the depth values used are from the stereo depth computed.
    \item \textbf{NIR}: 4 NIR wavelengths collected with \textit{NIR} camera
    \item \textbf{SWIR}: 7 wavelengths collected with \textit{SWIR} camera
\end{itemize}

It is to be noted that, the multi-spectral modalities \textit{NIR} and \textit{SWIR} might contain redundant channels, however, this is not considered at this stage since the cost of adding or subtracting another wavelength in the \textit{NIR} and \textit{SWIR} is minimal, i.e., only the cost of LED illuminators is added. We treat the channels as "Blocks" which come from a specific device. This was done since the objective is to reduce/ select an optimal subset of devices.

\subsubsection{Additional baselines}

In addition to the MC-PixBiS Network, we have added two additional PAD baselines in this section. The objective is to identify the performance change with different channels when different models are used. We add one feature based baseline and another recent CNN based PAD method for this comparison. These baselines are listed below 

\begin{itemize}

\item \textit{MC-RDWT-Haralick-SVM}: This is the handcrafted feature baseline we have considered. This is a multi-channel extension of the RDWT-Haralick-SVM method in \cite{agarwal2017face}. In our setting, for the multi-channel images, each channel is divided into a  $4 \times 4$ grid first. Haralick \cite{haralick1979statistical} features from RDWT decompositions are extracted from each of these patches. The final feature vectors are obtained by concatenating features from all grids across all the channels. An SVM framework is used together with these features for the PAD task.

\item \textit{MC-CDCN}: This is the multi-channel variant of the method proposed in \cite{yu2020multi}. This approach won first place in Multi-Modal Track in the ChaLearn Face Anti-spoofing Attack Detection Challenge@CVPR2020 \cite{liu2021cross}. This approach is essentially an extension of their previous work on central difference convolutional (CDC) networks to multiple channels. The core idea in CDC is the aggregation of the center-oriented gradient in local regions, the CDC is further combined with Vanilla convolutions. In the multi-modal setting, they extend the number of branches with CDC networks, following a late fusion strategy using non-shared component branches. We adapted the model to accept a varying number of input channels by introducing additional input branches. However, model complexity and computations requirements increase greatly as the number of channels increases due to non-shared branches.

\end{itemize}

We have performed experiments with individual channels as well as with the combination of channels in the ``grandtest-c'' protocol. We could not perform experiments involving MC-CDCN in some channel combinations due to the huge increase in parameter and computational complexity with the increase in the number of input channels. The change in number of parameters and computations are shown in Table. \ref{tab:parameters_complexity}. This also shows a practical advantage of fusing channels at input level as the computational and parameter increase are very minimal. 

The experimental results with three methods are shown in Table. \ref{tab:ablation_chanels_baselines}, in all three baselines the SWIR channel achieves the best individual performance, followed by RGB channel. This trend is common for both CNN based and feature based baselines. From the channel combinations models involving RGB and SWIR channel obtains the best results for both MC-PixBiS and Haralick-SVM baselines. In general, CNN based methods are outperforming feature based baseline in all the combinations. We have used the MC-PixBiS model for the further experiments as it obtains good results with a minimal parameter and computational complexity.

\begin{table}[ht]
\caption{Comparison of number of parameters and compute for the compared CNN models. For MC-CDCN the number of parameters and compute increases greatly as more 
channels are added.}

\begin{tabular}{c|r|r|r|r}
\toprule
\multicolumn{1}{c|}{\multirow{2}{*}{Channels}} & \multicolumn{2}{c|}{MC-PixBiS}                                 & \multicolumn{2}{c}{MC-CDCN}                                   \\ \cmidrule(l){2-5} 
\multicolumn{1}{c|}{}                          & \multicolumn{1}{c|}{Compute} & \multicolumn{1}{c|}{Parameters} & \multicolumn{1}{c|}{Compute} & \multicolumn{1}{c}{Parameters} \\ \midrule
1  &  4.52 GMac &  3.19 M &   47.48 GMac &   2.32 M \\
2  &  4.58 GMac &  3.19 M &   94.96 GMac &   4.64 M \\
3  &  4.64 GMac &  3.20 M &  142.44 GMac &   6.95 M \\
4  &  4.70 GMac &  3.20 M &  189.91 GMac &   9.27 M \\
5  &  4.76 GMac &  3.21 M &  237.39 GMac &  11.59 M \\
6  &  4.81 GMac &  3.21 M &  284.87 GMac &  13.90 M \\
7  &  4.87 GMac &  3.22 M &  332.35 GMac &  16.22 M \\
8  &  4.93 GMac &  3.22 M &  379.83 GMac &  18.54 M \\
9  &  4.99 GMac &  3.23 M &  427.30 GMac &  20.85 M \\
10 &  5.05 GMac &  3.23 M &  474.78 GMac &  23.17 M \\
11 &  5.11 GMac &  3.24 M &  522.26 GMac &  25.49 M \\
12 &  5.17 GMac &  3.24 M &  569.74 GMac &  27.80 M \\
13 &  5.23 GMac &  3.25 M &  617.21 GMac &  30.12 M \\ 

\bottomrule
\end{tabular}

\label{tab:parameters_complexity}
\end{table}

\begin{table*}[ht!]
\caption{Performance of three different models in the \textit{Grandtest-c} protocol in \textit{HQ-WMCA}. ACER in the \textit{test} set corresponding to BPCER 1\% threshold in \textit{validation} set is shown in the table.}
\centering

\begin{tabular}{lrrr}
\toprule

Channels &  \multicolumn{1}{c} {MC-PixBiS} &  \multicolumn{1}{c} {Haralick-SVM} &  \multicolumn{1}{c} {MC-CDCN}  \\
\midrule
              RGB  &         4.6     & 16.1    & 12.7     \\
                D  &         26.7    & 35.2    & 36.9     \\
                 T &        44.9     & 50.0    & 19.1     \\
              NIR  &         9.7     & 24.3    & 21.3     \\ 
     \textbf{SWIR} &         4.1     & 5.8     & 1.6      \\ \midrule
 \textbf{RGB-SWIR} &         0.3     & 2.5     & -        \\
      RGB-D-T-SWIR &         0.0     & 1.8     & -        \\
  RGB-D-T-NIR-SWIR &         0.0     & 1.5     & -        \\
           RGB-NIR &         0.7     & 11.6    & 6.0      \\
          NIR-SWIR &         2.7     & 4.3     & -        \\
             RGB-D &         6.4     & 18.8    & 11.2     \\
       RGB-D-T-NIR &         3.4     & 11.0    & 5.8      \\
             RGB-T &         6.9     & 10.4    & 11.0     \\
           RGB-D-T &         6.2     & 13.4    & 9.3      \\
               D-T &        17.0     & 26.2    & 16.4     \\
\bottomrule
\end{tabular}

\label{tab:ablation_chanels_baselines}
\end{table*}

In each of the combinations, we carried out model training and evaluation on all the protocols. The results in each of the protocols are tabulated in Table \ref{tab:ablation_chanels}.

\subsubsection{Results in \textit{Grandtest-c} protocol }

The \textit{Grandtest-c} protocol emulates the performance of a system under a wide variety of attacks, in a known attack scenario.

Out of individual channels (Table \ref{tab:ablation_chanels}), the \textit{SWIR} channel performs the best closely followed by the \textit{RGB} channel. Both these channels could detect the attacks very well even in the presence of a wide variety of attacks.

Not surprisingly, the combination of \textit{RGB-SWIR} achieves very good results with an ACER of 0.3 \%, which is better than their independent error rates, indicating the complementary nature of the information gained by the combination. The combinations \textit{RGB-D-T-NIR-SWIR} and \textit{RGB-D-T-SWIR} achieve perfect separation, however, these are supersets of \textit{RGB-SWIR} combination and the addition of other channels does not contribute much to the performance overall. Another combination that fares well is \textit{RGB-NIR}, which achieves a notable ACER of 0.7\%.

\begin{figure*}[ht!]
\centering
\includegraphics[width=0.99\textwidth]{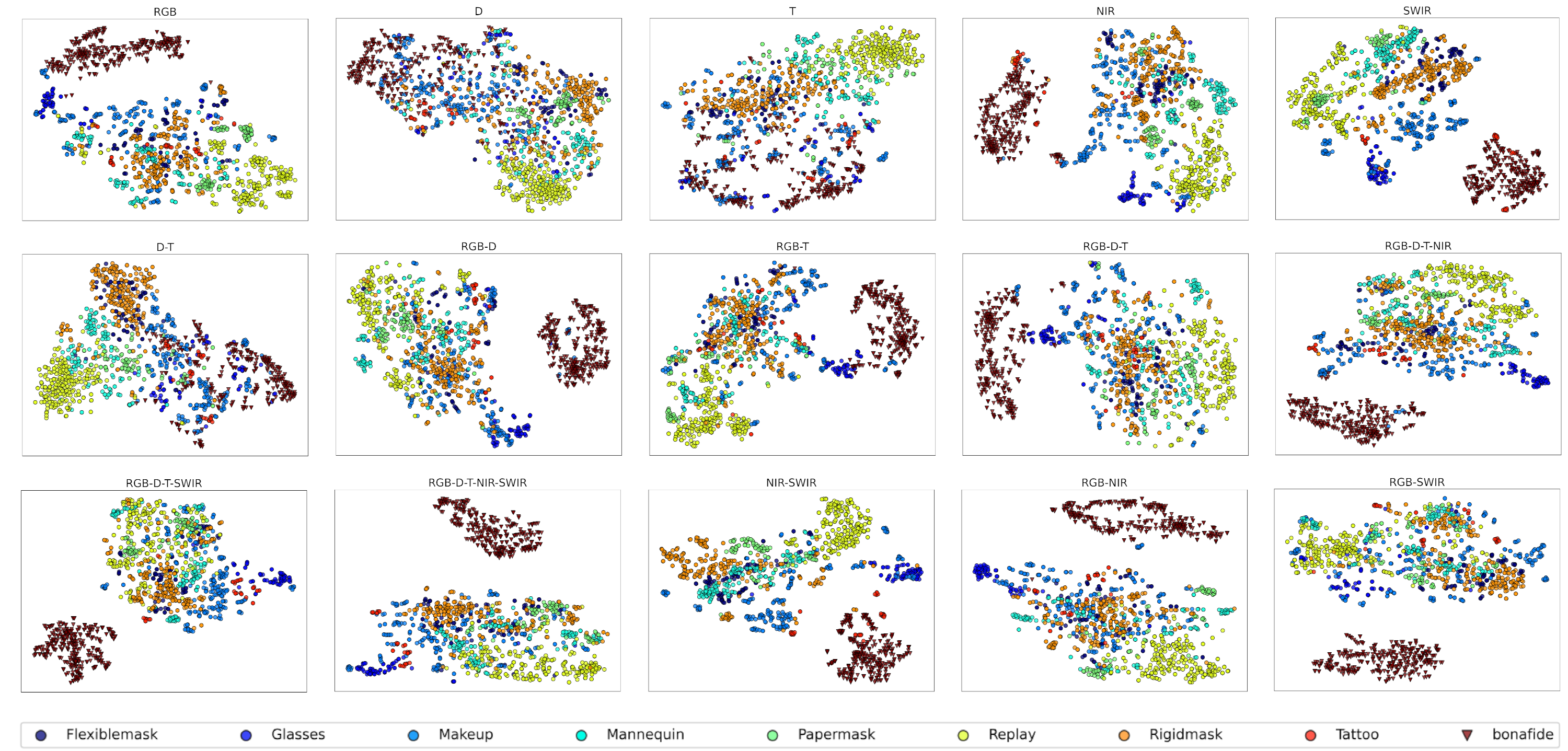}
\caption{t-SNE plots corresponding to different combinations of channels in the \textit{Grandtest-c} protocol. The first row shows the individual channels, second and third rows shows different combinations of channels (best viewed in color). }
\label{fig:tSNE}
\end{figure*}

The t-SNE plots for different combinations of channels is shown in Fig. \ref{fig:tSNE}. From the plots, it can be seen that combining different channels improves the separation between bonafide and attack samples.

\subsubsection{Results in \textit{Impersonation-c} protocol }

The \textit{Impersonation-c} protocol mostly consists of attacks aimed at impersonating another subject. From the experimental results, it can be seen that this is by far the easiest protocol among the three protocols considered.

\textit{NIR} channel performs the remarkably well followed by the \textit{RGB} channel. However, it is to be noted that most of the combinations of channels performed reasonably well in this protocol.

Consequently, the \textit{RGB-NIR} combination achieves perfect performance in the impersonation protocol. Several other combinations also perform reasonably well in this protocol.

\subsubsection{Results in \textit{Obfuscation-c} protocol }

The \textit{Obfuscation-c} protocol emulates the detection of obfuscation attacks such as makeups, glasses, and tattoos. This is by far the most difficult protocol among the three protocols considered. Most of the cases,  the attacks are partial and appear only in a part of the face.  This makes it harder to detect these attacks in general.

In this protocol, the \textit{SWIR} channel performs better compared to other channels, followed by the \textit{RGB} channel. Most of the other channels perform poorly. The success of the SWIR channel could be due to the specific spectral properties of skin, as compared to other PAIs.

There is a quick jump in performance when channels are used together. In fact, the \textit{RGB-SWIR} combination achieves an ACER of 0.7 \% in this challenging protocol, indicating the complementary nature of these channels. Another notable combination is \textit{RGB-NIR} with an ACER of 7.4 \%.

\subsubsection{Summary of channel-wise study}

Among the three different protocols with different difficulties and challenges, out of individual channels, \textit{SWIR} channels seem to perform the best. However, interestingly, it is followed closely by \textit{RGB}. The usefulness of these channels is visible in the combinations as well, with \textit{RGB-SWIR} achieving the best average results across the three protocols. This also indicates the complementary nature of discriminative information present in these channels for PAD. Another notable combination is \textit{RGB-NIR} which achieves an average error rate of 2.7\%. We have experimented with the depth coming from Intel Realsense (D-Intel) and the stereo depth computed (D-stereo), and have observed that the depth from stereo is slightly better in performance. The stereo camera obviates the requirement for an additional depth camera and hence we have used the depth coming from stereo for all the subsequent experiments. 


\subsection{Score Fusion Experiments}

\begin{table}[!b]
\caption{ACER in the \textit{test} set for various score fusion methods, feature fusion as compared to early fusion, in the \textit{Grandtest-c} protocol}
\centering
\resizebox{0.45\textwidth}{!}{

\begin{tabular}{crrrrrr}

\toprule

\multirow{2}{*}{Channels} & \multicolumn{4}{c}{Score fusion} & \multicolumn{1}{c}{\multirow{2}{*}{Feature fusion}} & \multicolumn{1}{c}{\multirow{2}{*}{Early fusion}} \\ \cmidrule{2-5}
                          &   \multicolumn{1}{c} {GMM} &   \multicolumn{1}{c} {LLR} &   \multicolumn{1}{c} {MLP} &  \multicolumn{1}{c} {MEAN}    & \multicolumn{1}{c}{}                              \\

\midrule
(RGB,SWIR)            &   4.4 &   0.8 &   0.7 &   0.7          &4.1  &  \textbf{0.3} \\
(RGB,D,T,SWIR)        &   4.3 &   3.3 &   4.0 &   3.6          &6.4  &  \textbf{0.0} \\
(RGB,D,T,NIR,SWIR)    &   4.0 &   4.1 &   5.1 &   4.1          &6.9  &  \textbf{0.0} \\
(RGB,NIR)             &   3.9 &   4.9 &   9.0 &   4.9          &4.1  &       \textbf{0.7} \\
(NIR,SWIR)            &   7.7 &   2.7 &   4.7 &   \textbf{2.1} &6.1  &  2.7 \\
(RGB,D)               &   4.6 &   \textbf{3.4} &   4.2 &   4.2 &4.9  &  6.4 \\
(RGB,D,T,NIR)         &   7.1 &   6.7 &   6.6 &   6.6          &7.1  &  \textbf{3.4} \\
(RGB,T)               &   \textbf{4.3} &   7.0 &   6.2 &   7.0 &6.7  &  6.9 \\
(RGB,D,T)             &   7.1 &   6.8 &   9.6 &   \textbf{6.2} &6.7  &  \textbf{6.2} \\
(D,T)                 &  43.2 &  18.4 &  18.4 &  18.4          &41.0 & \textbf{17.0} \\

\bottomrule
\end{tabular}

}

\label{tab:fusion}

\end{table}

\begin{figure*}[!t]
\centering
\includegraphics[width=0.75\textwidth]{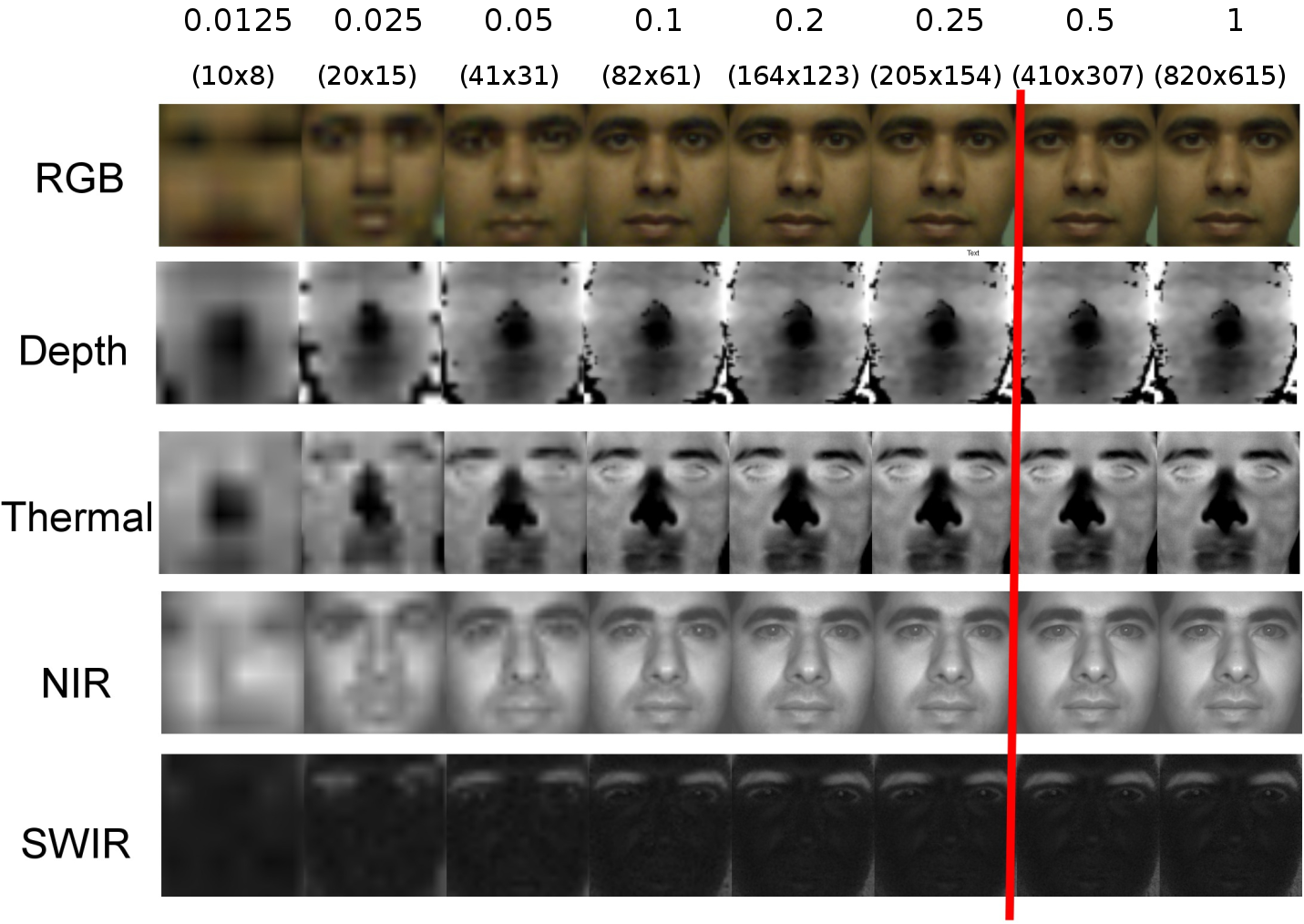}
\caption{Preprocessed images with various scaling factors in RGB, Thermal, Depth, NIR and SWIR; upto a scaling of 0.3 the image quality is not affected since the original face size is larger than the resized size of $224 \times 224$. The approximate resolution of face region after scaling is also shown. }
\label{fig:images_caling}
\end{figure*}

So far in the channel selection experiments, for combining channels we have stacked them at the input stage. This could have some disadvantages in case a channel is faulty or missing at test time. Also, the stacked model could have more parameters and could be more prone to over-fitting, while trained with a small dataset.

The objective here is to evaluate the performance of various models with score-level fusion and feature-level fusion and to contrast the performance as compared to model-level. Instead of training CNNs with stacked channels, we performed fusion experiments on CNN models trained on individual channels separately. For score-level fusion, scores from individual systems are used as features to train a fusion model. For feature fusion, features extracted from each individual model are concatenated and combined with SVM to obtain the final scores. This allows redundancy if a channel is missing at the deployment stage.

We used four different fusion models trained on top of the scalar scores returned by the component CNN's. The score fusion models used are:

\begin{itemize}
\item \textbf{GMM}: Gaussian Mixture Model
\item \textbf{Mean}: Simple mean of different model outputs
\item \textbf{LLR}: Linear Logistic Regression
\item \textbf{MLP}: Multi-layer Perceptron
\end{itemize}

The results from the score fusion, and feature fusion models on the \textit{Grandtest-c} protocol are summarized in Table. \ref{tab:fusion}. The channels column shows the channels corresponding to individual CNN's used in the model, and the performance with different fusion models are tabulated. From the results and comparing the results of score fusion, feature fusion and early fusion with the same combination of channels (as in Table \ref{tab:ablation_chanels}), it can be seen that in most of the cases, early fusion performs better except for \textit{NIR-SWIR}, \textit{RGB-D} and \textit{RGB-T} combinations where score fusion performs better.

This shows that score level fusion and feature level fusion fails to capture inter-channel dependencies in many cases which could be done with an early fusion approach. Nevertheless, the fusion method is still interesting from a deployment point of view.


\subsection{Evaluation on Varying Image Resolutions}

\begin{table*}[h]
\caption{ACER in the \textit{eval} for various scaling values in the \textit{Grandtest-c} protocol in \textit{HQ-WMCA}, with reprojection and unit spectral normalization.}
\centering
\begin{tabular}{lRRRRRRRRR}
\toprule
Channels   &  \multicolumn{1}{c} {0.0125} &  \multicolumn{1}{c} {0.025} &  \multicolumn{1}{c} {0.05} &   \multicolumn{1}{c} {0.1} &   \multicolumn{1}{c} {0.2} &  \multicolumn{1}{c} {0.25} &   \multicolumn{1}{c} {0.5} &   \multicolumn{1}{c} {1.0} \\ \midrule
RGB              &    16.9 &   10.4 &   5.5 &   5.3 &   \textbf{3.2} &   9.1 &   \textbf{3.2} &   4.6 \\
D                &    38.8 &   36.4 &  26.0 &  24.8 &  23.6 &  25.4 &  \textbf{22.4} &  26.7 \\
T                &    32.9 &   24.8 &  \textbf{23.2} &  42.7 &  34.6 &  33.6 &  31.5 &  44.9 \\
NIR              &    24.5 &   21.1 &  30.2 &  23.7 &  17.6 &  16.6 &  13.2 &   \textbf{9.7} \\
SWIR             &     4.1 &    1.8 &   1.7 &   1.6 &   \textbf{1.5} &   3.0 &   3.2 &   4.1 \\ \midrule
RGB-T            &    16.7 &   11.7 &   7.2 &  11.6 &   8.9 &   \textbf{4.2} &   \textbf{4.2} &   6.9 \\
RGB-D-T          &    20.4 &    8.1 &   7.5 &   \textbf{5.1} &   8.2 &   9.1 &   9.6 &   6.2 \\
D-T              &    25.1 &   28.9 &  24.6 &  18.6 &  26.3 &  33.2 &  45.8 &  \textbf{17.0} \\
RGB-D            &    26.2 &    8.1 &   5.2 &   6.7 &   \textbf{3.4} &  10.2 &   4.8 &   6.4 \\
RGB-D-T-NIR      &    10.4 &   11.3 &   6.7 &   5.1 &   5.3 &   5.2 &   7.6 &   \textbf{3.4} \\
RGB-D-T-NIR-SWIR &     1.2 &    0.3 &  \textbf{0.0} &   0.3 &   \textbf{0.0} &   \textbf{0.0} &   \textbf{0.0} &   \textbf{0.0} \\
RGB-D-T-SWIR     &     1.1 &    0.1 &   \textbf{0.0} &   0.1 &   \textbf{0.0} &   \textbf{0.0} &   \textbf{0.0} &   \textbf{0.0} \\
NIR-SWIR         &     5.9 &    3.2 &   1.9 &   2.2 &   \textbf{1.2} &   3.7 &   3.1 &   2.7 \\
RGB-NIR          &    11.2 &    5.0 &   2.5 &   0.6 &   2.7 &   1.0 &   \textbf{0.5} &   0.7 \\
RGB-SWIR         &     1.1 &    0.6 &   \textbf{0.0} &   0.5 &   0.1 &   \textbf{0.0} &   \textbf{0.0} &   0.3 \\
\bottomrule
\end{tabular}

\label{tab:res_grandtest}

\end{table*}

\begin{table*}[h]
\caption{ACER in the \textit{eval} for various scaling values in the \textit{Impersonation-c} protocol in \textit{HQ-WMCA}, with reprojection and unit spectral normalization.}
\centering
\begin{tabular}{lRRRRRRRRR}
\toprule
Channels   &  \multicolumn{1}{c} {0.0125} &  \multicolumn{1}{c} {0.025} &  \multicolumn{1}{c} {0.05} &   \multicolumn{1}{c} {0.1} &   \multicolumn{1}{c} {0.2} &  \multicolumn{1}{c} {0.25} &   \multicolumn{1}{c} {0.5} &   \multicolumn{1}{c} {1.0} \\ \midrule
RGB              &     6.7 &    1.8 &   0.8 &  1.4 &  \textbf{0.0} &   1.1 &  0.4 &  0.6 \\
D                &    23.1 &    9.1 &   5.9 &  5.3 &  4.5 &   4.1 &  4.1 &  \textbf{3.8} \\
T                &     3.8 &    1.7 &   2.7 &  2.1 &  1.3 &   1.2 &  \textbf{0.9} &  \textbf{0.9} \\
NIR              &    22.7 &    8.2 &   0.5 &  0.4 &  0.3 &   0.3 &  \textbf{0.0} &  0.1 \\
SWIR             &     0.1 &    \textbf{0.0} &   0.6 &  0.2 &  1.1 &   0.2 &  0.4 &  1.8 \\ \midrule
RGB-T            &     2.1 &    1.7 &   4.3 &  4.3 &  \textbf{0.9} &   2.5 &  3.1 &  2.5 \\
D-T              &     2.9 &    \textbf{1.8} &   3.6 &  6.1 &  2.8 &   1.9 &  2.2 &  4.1 \\
RGB-D            &     5.7 &    2.8 &   2.0 &  2.0 &  1.9 &   \textbf{0.5} &  1.3 &  1.7 \\
RGB-D-T-NIR      &     1.1 &    0.5 &   0.1 &  \textbf{0.0} &  0.5 &   0.1 &  \textbf{0.0} &  0.1 \\
RGB-D-T          &     \textbf{1.0} &    4.2 &   3.8 &  4.3 &  5.6 &   4.3 &  2.6 &  3.4 \\
RGB-NIR          &     0.8 &    0.5 &   \textbf{0.0} &  0.2 &  0.1 &   \textbf{0.0} &  0.4 &  \textbf{0.0} \\
RGB-D-T-NIR-SWIR &     0.3 &    0.4 &   \textbf{0.1} &  2.2 &  0.5 &   0.3 &  0.6 &  0.7 \\
NIR-SWIR         &     0.1 &    0.1 &   0.4 &  0.2 &  \textbf{0.0} &   1.5 &  0.9 &  0.4 \\
RGB-SWIR         &     \textbf{0.1} &    0.7 &   1.3 &  2.4 &  1.4 &   2.0 &  0.7 &  2.0 \\
RGB-D-T-SWIR     &     0.2 &    0.4 &   0.1 &  0.4 &  0.9 &   \textbf{0.0} &  1.9 &  2.5 \\

\bottomrule
\end{tabular}

\label{tab:res_impersonation}

\end{table*}

\begin{table*}[h]
\caption{ACER in the \textit{test} for various scaling values in the \textit{Obfuscation-c} protocol in \textit{HQ-WMCA}, with reprojection and unit spectral normalization.}
\centering
\begin{tabular}{lRRRRRRRRR}
\toprule
Channels   &  \multicolumn{1}{c} {0.0125} &  \multicolumn{1}{c} {0.025} &  \multicolumn{1}{c} {0.05} &   \multicolumn{1}{c} {0.1} &   \multicolumn{1}{c} {0.2} &  \multicolumn{1}{c} {0.25} &   \multicolumn{1}{c} {0.5} &   \multicolumn{1}{c} {1.0} \\ \midrule
RGB              &    28.2 &   21.3 &  16.6 &  20.3 &  16.7 &  \textbf{12.2} &  12.5 &  14.8 \\
D                &    47.3 &   48.9 &  46.0 &  48.2 &  \textbf{45.1} &  50.0 &  47.9 &  48.7 \\
T                &    48.6 &   49.7 &  49.9 &  49.6 &  \textbf{48.2} &  49.8 &  49.9 &  50.0 \\
NIR              &    45.5 &   41.1 &  41.5 &  39.6 &  \textbf{34.6} &  42.3 &  40.6 &  39.6 \\
SWIR             &    11.6 &    5.6 &   6.1 &   5.6 &   \textbf{4.7} &   6.5 &   7.1 &   9.2 \\ \midrule
RGB-D            &    30.3 &   20.5 &  21.4 &  20.6 &  16.7 &  20.0 &  23.6 &  \textbf{12.4} \\
RGB-T            &    35.7 &   26.7 &  19.4 &  20.0 &  \textbf{16.4} &  22.6 &  23.2 &  17.6 \\
D-T              &    \textbf{48.1} &   49.5 &  50.0 &  50.0 &  49.9 &  48.8 &  50.0 &  49.9 \\
RGB-D-T          &    35.2 &   31.2 &  22.8 &  18.2 &  \textbf{18.0} &  25.8 &  25.6 &  20.1 \\
RGB-D-T-SWIR     &     9.0 &    1.7 &   4.8 &   0.1 &   2.1 &   \textbf{0.0} &   \textbf{0.0} &   \textbf{0.0} \\
RGB-D-T-NIR      &    28.9 &   26.4 &  18.7 &  \textbf{13.9} &  19.9 &  17.8 &  19.9 &  17.4 \\
RGB-SWIR         &     3.7 &    0.3 &   \textbf{0.0} &   \textbf{0.0} &   \textbf{0.0} &   \textbf{0.0} &   0.1 &   \textbf{0.0} \\
RGB-NIR          &    21.5 &   18.5 &  16.3 &  12.1 &  10.1 &  13.7 &  12.2 &   \textbf{7.4} \\
NIR-SWIR         &     7.2 &    8.8 &   \textbf{6.7} &   7.9 &  10.1 &  12.5 &   7.9 &   8.4 \\
RGB-D-T-NIR-SWIR &     3.0 &    8.4 &   0.5 &   0.5 &   0.2 &   \textbf{0.0} &   \textbf{0.0} &   2.3 \\

\bottomrule
\end{tabular}

\label{tab:res_obfuscation}

\end{table*}

This subsection discusses the variation of the performance with respect to the image resolution of different imaging modalities. To emulate a lower resolution sensor, we have systematically down-sampled the source image to a lower resolution followed by scaling as required by the framework.

The average face size in the \textit{RGB} channel in \textit{HQ-WMCA} dataset was $819.5 (\pm 94.3) \times 614.1 (\pm 63.9)$ pixels. The MC-PixBiS model requires an input of size $224 \times 224$, and hence the resolution of the original image is much higher than what is required for the framework. This also means that we can safely scale the images up to a scaling factor close to $0.3$ without causing much degradation in the preprocessed files. For different scaling factors, the source image (not just the face part) is down-sampled with the scaling factor first to emulate a low-resolution sensor. The rest of the processing is performed with this scaled image, i.e., face detection and resizing of cropped face region to $224 \times 224$ image in the scaled raw image. In summary, the raw image undergoes two scaling steps, i.e., first to emulate a low-resolution sensor and then resizing the cropped face region. This introduces some minor artifacts, due to the interpolation stages (even with scale factors greater than 0.3). The image quality after preprocessing, with different scaling factors, is shown in Fig. \ref{fig:images_caling}. All the channels are first aligned to the \textit{RGB} channel spatially, and the scaling is applied uniformly to all the aligned channels.

The \textit{SWIR} channel is the most expensive sensor, by far from other acquisition channels in the sensor suite. Typically available sensor resolutions for \textit{SWIR} are, $640 \times 512$,  $320 \times  256$, $128 \times 128$ and $64 \times 64$ (based on availability in market), approximately corresponding to the scale factors of 1, 0.5, 0.2, and 0.1 respectively in this study. We have performed the experiments in all the protocols with different scaling factors and the results are tabulated in Table \ref{tab:res_grandtest}, \ref{tab:res_impersonation} and \ref{tab:res_obfuscation}.

From the results, it can be seen that the performance degradation occurs only at very low resolutions. We observed that the performance starts degrading greatly at a very low scaling factor like $0.0125$. 

Surprisingly, in some cases, it can be seen that the performance even gets better at lower resolutions. This can be due to the removal of high-frequency signals, which may not be relevant in the specific scenarios. In some cases, the spectra are more important compared to the spatial information, and in such scenarios, the performance improves when the resolution is low so that some amount of over-fitting which could have been occurred in original resolution goes away. One way to use this information is to use blurring as a data augmentation strategy while training future models with these channels.

Another interesting point to note in the tables is that, in all the protocols, the \textit{SWIR} channel performance is largely unaffected by the scaling. The performance degrades only at very low resolutions. This is remarkable since the \textit{SWIR} sensor is the most costly sensor among the sensors in the hardware suite. If the \textit{SWIR} channel could operate at a lower resolution, this would significantly reduce the cost of the entire sensor suite. One reason for this performance is the spectral nature of the SWIR channel, as compared to the spatial nature of other channels. The spectral nature of SWIR channel could be important for identifying skin pixels. In such a scenario,  a very low resolution such as $64 \times 64$ could be enough to achieve the desired performance. This is a very important result since it reduces the cost of the entire sensor suite by a significant amount. Besides, the combination of \textit{RGB-SWIR} performs best overall.


\subsection{Unseen Attack Experiments}

So far, all the experiments considered known attack scenarios. In this set of experiments, we evaluate the robustness of the PAD systems in unseen attack scenarios. Specifically, different sub-protocols were created, which systematically exclude one specific attack in the training and validation sets. The \textit{test} set consists of bonafide samples and the attack which was left out of the training set. This emulates actually encountering an unseen attack in a real-world environment. The performance of various combinations of channels is shown in Table \ref{tab:unseen}.

\begin{table*}[h]
\caption{ACER in the \textit{test} for various unseen attack protocols, with reprojection based alignment and unit normalization,  with original resolution}
\centering
\resizebox{0.9\textwidth}{!}{
\begin{tabular}{lRRRRRRRRRR}
\toprule
\multicolumn{1}{l} {Channels} &\multicolumn{1}{l} {FlexMask} &\multicolumn{1}{l} {Glasses} &\multicolumn{1}{l} {Makeup} &\multicolumn{1}{l} {Mannequin} &\multicolumn{1}{l} {Papermask} &\multicolumn{1}{l} {Rigidmask} &\multicolumn{1}{l} {Tattoo} &\multicolumn{1}{l} {Replay} &\multicolumn{1}{l} {Mean} &\multicolumn{1}{l} {Std} \\
\midrule
RGB               &         11.0        & 49.3           & 32.3          & 0.0        &  0.5        & 28.2            &   46.4            &  2.3            &21.2 & 20.4 \\
D                 &         41.6        & 49.1           & 50.1          &34.8        & 18.1        & 49.7            &   41.3            &  8.6            &36.6 & 15.5 \\
T                 &         47.3        & 50.3           & 49.9          &28.2        & 50.0        &  3.1            &   50.0            &  0.4            &34.9 & 21.7 \\
NIR               &         26.6        & 48.3           & 46.5          & 0.4        &\textbf{0.0} & 15.7            &   41.0            &  3.0            &22.6 & 20.7 \\
SWIR              & \textbf{0.0}        & 44.6           & 38.6          &\textbf{0.0}&\textbf{0.0} &  3.0            &   47.3            & \textbf{0.0}    &\textbf{16.6} & 22.3 \\ \midrule
RGB-D-T-NIR       &          4.2        & 50.0           & 41.7          & 0.0        &  0.0        & 33.3            &   38.0            &  0.0            &20.9 & 21.7 \\
RGB-D-T           &         28.3        & 50.0           & 33.8          & 3.7        &  0.0        & 40.2            &   41.6            &  6.5            &25.5 & 19.4 \\
RGB-D-T-SWIR      &          0.4        &  6.1           & 26.5          & 0.0        &  0.0        &\textbf{0.3}     &   41.8            &  0.0            & 9.3 & 15.9 \\
D-T               &         26.5        & 50.0           & 50.5          &46.2        & 44.2        & 33.7            &   50.0            & 41.4            &42.8 & 8.6  \\
RGB-D-T-NIR-SWIR  &          0.6        &  1.4           & 28.4          & 0.0        &  0.0        &  0.4            &   44.9            &  0.0            & 9.4 & 17.3 \\
RGB-T             &         25.5        & 50.0           & 34.8          &15.8        &  0.0        &  1.7            &   32.9            &  0.2            &20.1 & 18.7 \\
NIR-SWIR          &          0.0        & 47.2           & 31.6          & 0.0        &  0.0        &  1.7            &   50.0            &  0.0            &16.3 & 22.6 \\
RGB-D             &         12.7        & 47.4           & 29.1          & 0.9        &  0.0        & 42.1            &   20.8            &  4.1            &19.6 & 18.4 \\
RGB-NIR           &         13.5        & 41.4           & 46.2          & 0.6        &  0.1        & 29.6            &\textbf{2.4}       &  0.1            &16.7 & 19.5 \\
\textbf{RGB-SWIR} &          0.3        & \textbf{0.5}   & \textbf{23.1} & 0.8        &  0.6        &  1.2            &   27.4            &  1.3            & \textbf{6.9} & 11.3 \\
\bottomrule
\end{tabular}
} 
\label{tab:unseen}
\end{table*}

From Table \ref{tab:unseen}, it can be seen that out of individual channels, the \textit{SWIR} channel achieves the best individual average ACER. For the channel combinations, \textit{RGB-SWIR} achieves the best performance with an average ACER of 6.9\%. Among the different attacks, makeups, tattoos, and glasses appears to be the most difficult attacks to detect if they are not seen in the training set. The \textit{RGB-NIR} model works best in detecting tattoos. Surprisingly, the \textit{RGB-SWIR} combination achieved reasonable performance in both seen and unseen attack protocols.

\begin{figure}[ht!]
\centering
\includegraphics[width=0.48\textwidth]{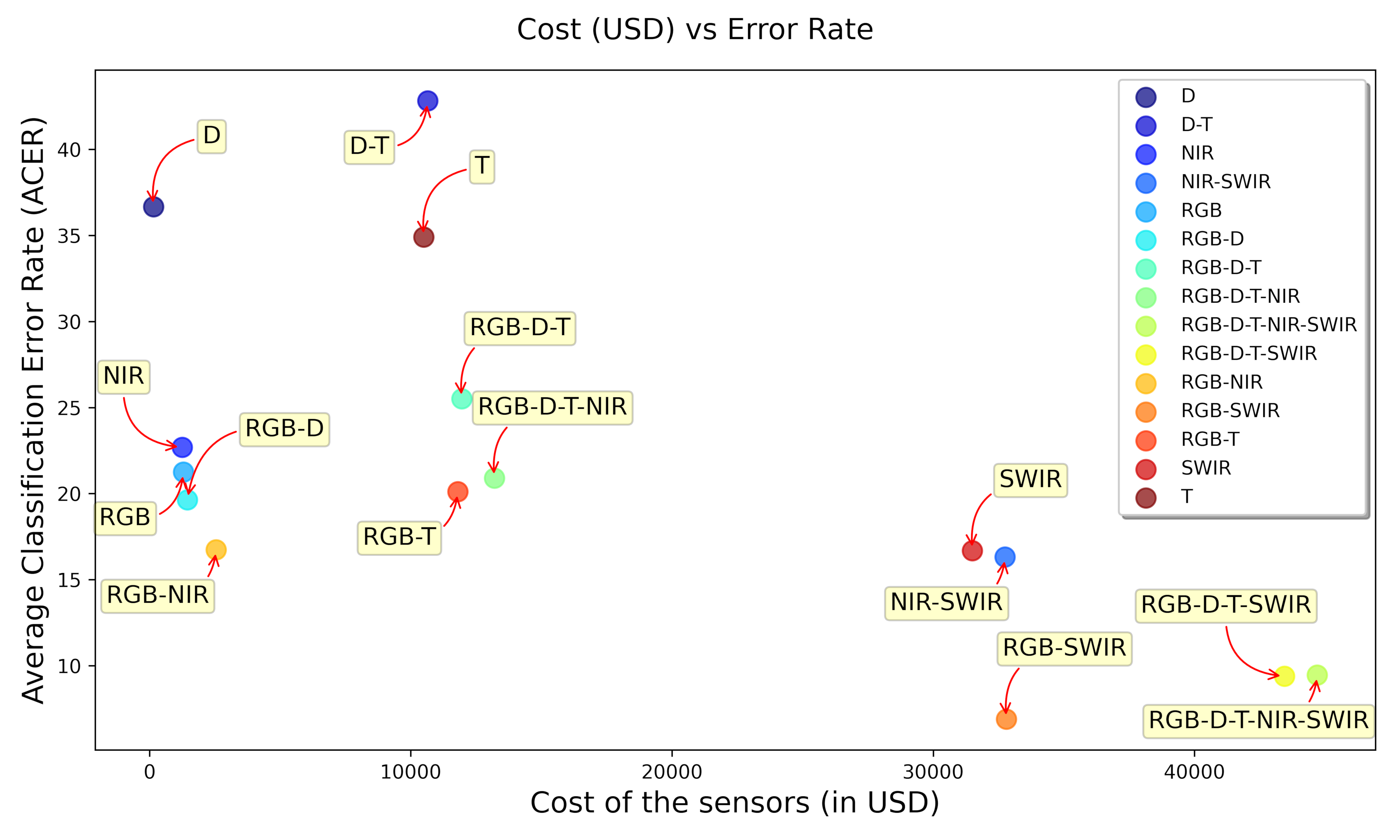}
\caption{Cost of the hardware vs. average ACER in unseen attack protocol, cost and performance was calculated based on the original sensor resolutions as available with the dataset.}
\label{fig:cost_vs_perf}
\end{figure}

We have also added a figure (Fig. \ref{fig:cost_vs_perf}) showing the performance in terms of average ACER in the unseen attack protocols against the cost of the sensors used. The cost and performance were calculated based on the original sensor resolutions as available with the dataset. However, the cost would further reduce with lower resolution sensors. 

\subsection{Performance with different Wavelengths in NIR and SWIR}
The channel wise study in section \ref{sec:channel_exp} aimed to identify the best channels as groups. However based on the interesting results, we investigated the importance of different wavelengths in different SWIR and NIR channels. Specifically, since \textit{RGB-SWIR} and \textit{RGB-NIR} were performing well, we conducted additional experiments with different combinations of \textit{RGB} and individual wavelengths from \textit{NIR} and \textit{SWIR}, one at a time. The results from the unseen attack protocols are tabulated in Table \ref{loo:subnir_swir}. Surprisingly, the \textit{RGB\_SWIR\_1450nm} alone performs comparable to or better than \textit{RGB-SWIR} combination. Indeed, the $1450 nm$ is closer to water absorption frequency \cite{wilson2015review}, which is characteristic to human skin \cite{nicolo2012long} which could explain the robustness of this frequency. This wavelength was also observed to be important in previous studies as well \cite{heusch2020deep}.

\begin{table*}[h]
\caption{ACER in the \textit{test} for various unseen attack protocols, with reprojection based alignment and unit normalization,  with original resolution}
\centering
\resizebox{0.9\textwidth}{!}{
\begin{tabular}{lRRRRRRRRRR}
\toprule
\multicolumn{1}{l} {Channels} &\multicolumn{1}{l} {FlexMask} &\multicolumn{1}{l} {Glasses} &\multicolumn{1}{l} {Makeup} &\multicolumn{1}{l} {Mannequin} &\multicolumn{1}{l} {Papermask} &\multicolumn{1}{l} {Rigidmask} &\multicolumn{1}{l} {Tattoo} &\multicolumn{1}{l} {Replay} &\multicolumn{1}{l} {Mean} &\multicolumn{1}{l} {Std} \\
\midrule

RGB              & 11.0 & 49.3 & 32.3 & 0.0 &  0.5 & 28.2 &   46.4 &  2.3  &21.2 & 20.4 \\ \midrule
\textbf{RGB\_NIR\_735nm}    &          10.1 &     42.0 &    39.7 &        0.0 &        0.0 &       26.6 &     4.5 &     0.0   & \textbf{15.3}  & 18.0\\
RGB\_NIR\_850nm             &          13.2 &     45.6 &    33.0 &        0.2 &        0.0 &        4.1 &    46.7 &     7.2   & 18.7  & 19.9\\
RGB\_NIR\_940nm             &           0.4 &     49.3 &    43.2 &        0.7 &        0.0 &       28.3 &    13.6 &     4.1   & 17.4  & 20.2\\
RGB\_NIR\_1050nm            &          22.3 &     50.0 &    29.2 &        0.9 &        0.0 &       31.8 &    36.1 &     0.3   & 21.3  & 19.0\\ \midrule
RGB\_SWIR\_940nm            &           8.1 &     48.8 &    32.6 &        0.3 &        0.6 &       21.5 &    32.4 &     7.0   & 18.9  & 17.7\\
RGB\_SWIR\_1050nm           &           0.0 &     14.5 &    29.7 &        0.0 &        0.0 &        1.7 &    49.8 &     0.0   & 11.9  & 18.6\\
RGB\_SWIR\_1200nm           &          24.0 &     18.0 &    24.1 &        0.3 &        1.0 &        9.8 &     1.8 &     0.0   &  9.8  & 10.6\\
RGB\_SWIR\_1300nm           &           0.0 &     42.5 &    29.5 &        0.0 &        0.0 &        0.1 &     4.9 &     0.0   &  9.6  & 16.7\\
\textbf{RGB\_SWIR\_1450nm}  &           0.1 &      4.2 &    25.2 &        0.3 &        0.5 &        0.0 &    24.1 &     0.0   &  \textbf{6.8}  & 11.1\\
RGB\_SWIR\_1550nm           &           0.6 &     13.4 &    26.8 &        1.2 &        0.3 &        0.2 &    37.9 &     0.5   & 10.1  & 14.7\\
RGB\_SWIR\_1650nm           &           0.3 &     14.7 &    29.5 &        0.0 &        0.4 &        0.0 &    35.2 &     0.4   & 10.01 & 14.7\\

\bottomrule
\end{tabular}
} 
\label{loo:subnir_swir}
\end{table*}

\begin{table}[h]
\caption{ACER in the \textit{test} for the main three protocols with sub channel performance}
\centering
\resizebox{0.49\textwidth}{!}{
\begin{tabular}{lRRRR}
\toprule

\multicolumn{1}{l} {Channels} &\multicolumn{1}{l} {Grandtest-c} &\multicolumn{1}{l} {Impersonation-c} &\multicolumn{1}{l} {Obfuscation-c} &\multicolumn{1}{l} {Mean} \\
\midrule

RGB                         &           4.6 &                    0.6 &                 14.8    & 6.6 \\
NIR                         &           9.7 &                    0.1 &                 39.6    & 16.4 \\
SWIR                        &           4.1 &                    1.8 &                  9.2    & \textbf{5.0} \\ \midrule
\textbf{RGB\_NIR\_735nm}    &           0.9 &                    0.0 &                  4.4    & \textbf{1.7} \\
RGB\_NIR\_850nm             &           4.7 &                    0.2 &                 10.6    & 5.1 \\
RGB\_NIR\_940nm             &           3.1 &                    0.0 &                 16.3    & 6.4 \\
RGB\_NIR\_1050nm            &           3.8 &                    0.0 &                 18.1    & 7.3 \\ \midrule
RGB\_SWIR\_940nm            &           1.2 &                    0.7 &                  8.4    & 3.4 \\
RGB\_SWIR\_1050nm           &           0.9 &                    1.9 &                  5.2    & 2.6 \\
RGB\_SWIR\_1200nm           &           1.2 &                    1.1 &                  7.5    & 3.2 \\
RGB\_SWIR\_1300nm           &           0.2 &                    0.9 &                  6.9    & 2.6 \\
RGB\_SWIR\_1450nm           &           0.2 &                    1.1 &                  1.0    & 0.7 \\
\textbf{RGB\_SWIR\_1550nm}  &           0.3 &                    0.6 &                  0.0    & \textbf{0.3} \\
RGB\_SWIR\_1650nm           &           0.5 &                    0.5 &                  0.8    & 0.6 \\

\bottomrule
\end{tabular}
} 
\label{all:subnir_swir}
\end{table}

Also, we have performed the same set of experiments with the known attack protocols as well, and the results are tabulated in Table. \ref{all:subnir_swir}. In this set of experiments, \textit{RGB\_SWIR\_1550nm} appears to perform the best on average, closely followed by \textit{RGB\_SWIR\_1650nm}
and \textit{RGB\_SWIR\_1450nm} \footnote{$SWIR_{w}$ and $NIR_{w}$, denote the images captured at a wavelength of $w$ nanometers}. And among \textit{NIR} channels, \textit{RGB\_NIR\_735nm} performs the best as with the unseen attack protocols.

\subsection{Discussions}

From the experiments, the findings from different analyses and overall observations are summarized in this section.

From the channel wise selection experiments, it was clear that combining different channels improves the performance. Even when the performance of individual channels is poor, the combinations of channels were found to improve the performance even in most challenging scenarios. In general, the SWIR channel was found to be very beneficial for a wide variety of scenarios, followed by the \textit{RGB} channel. Also, combining \textit{RGB} and \textit{SWIR} improved the performance greatly, and it could achieve significant performance improvement in most of the challenging scenarios. \textit{RGB} and \textit{NIR} combination also works very well in some cases.  As for the attacks in impersonation attack protocols, most of the channel seems to work well. Most of the methods fail to correctly identify obfuscation attacks, and they are indeed much harder to identify.

The fusion experiments performed show that using channels together in an early fusion strategy provides more accurate results compared to score fusion and feature fusion in general. In most cases, the performance degraded when the models were fused in score level or feature level as opposed to stacking at the input level.

Experiments with different image resolutions reveal important aspects useful for practical deployment scenarios. The accuracy of the models degraded only at very low resolutions, meaning that PAD systems could achieve competitive performance with a low-resolution \textit{SWIR} sensor in the hardware, since the cost of the sensor decreases a lot with a lower sensor resolution.

For reliable use of face recognition, the PAD modules used should be robust against unseen attacks. Hence the evaluation of various channels against unseen attacks is of specific importance from a security perspective. It was observed that the \textit{RGB-SWIR} model achieves remarkable performance in unseen attack protocol.

Intrigued by the success of \textit{SWIR}, and specifically \textit{RGB-SWIR} models, we performed further experiments to understand which wavelength is more informative. This experiment was performed in both \textit{NIR} and \textit{SWIR} channels (Table. \ref{all:subnir_swir}, and Table. \ref{loo:subnir_swir}). Interestingly, \textit{RGB\_SWIR\_1450nm} achieved comparable performance to \textit{RGB-SWIR} in both unseen and known attack protocol. Not surprisingly, the \textit{SWIR} wavelength \textit{1450nm} is closer to a characteristic feature of water absorption, which explains the robustness of this particular wavelength. Indeed, this makes the separation of skin from other attack instruments easier, even when the sensor resolution is low. One could opt for a low-resolution \textit{SWIR} camera with a high-resolution \textit{RGB} camera, keeping the cost low retaining the good performance.

In this study, we have focused on the analysis of channels and image resolutions. Further, this study can be extended to other image degradations like quantization of channels, the dynamic range of sensors, and the effect of noise. 


\section{Conclusions}
\label{sec:conclusion}
In this work, we analyze the usage of different channels in PAD and perform an extensive set of experiments in terms of channels used and the resolution requirements of channels. We have shown that the combination of \textit{RGB} and \textit{SWIR} leads to excellent performance in both seen and unseen attacks. In general, the \textit{SWIR} channel performs well in detecting challenging obfuscation attacks, and the \textit{NIR} channel performs well in detecting impersonation attacks.  The combination of \textit{RGB} and \textit{SWIR} shows excellent performance gain compared to the performance of their individual performance. Further, it was observed that the resolution requirements for the \textit{SWIR} channel can be very low and this paves the way for designing compact and cheap multi-channel systems for PAD in high-security applications.  We hope that the results from the experiments reported in this paper will act as pointers for sensor selection for secure face recognition applications. Though the experiments have been conducted on one architecture only, the source code to perform these experiments are made publicly available making it possible to reproduce and extend the work with new architectures.



\section*{Acknowledgments}
Part of this research is based upon work supported by the Office of the Director of National Intelligence (ODNI), Intelligence Advanced Research Projects Activity (IARPA), via IARPA R\&D Contract No. 2017-17020200005. The views and conclusions contained herein are those of the authors and should not be interpreted as necessarily representing the official policies or endorsements, either expressed or implied, of the ODNI, IARPA, or the U.S. Government. The U.S. Government is authorized to reproduce and distribute reprints for Governmental purposes notwithstanding any copyright annotation thereon.

The authors would like to thank Zohreh Mostaani for conducting the data collection campaign and the data curation. 

\ifCLASSOPTIONcaptionsoff
  \newpage
\fi

\bibliography{ref}
\bibliographystyle{IEEEtran}

\begin{IEEEbiography}
  [{\includegraphics[width=1in,height=1.25in,clip,keepaspectratio]{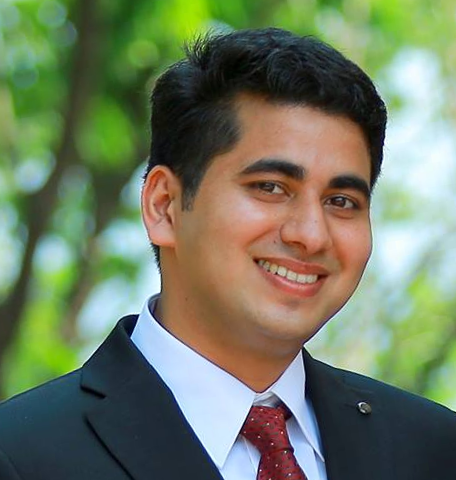}}]{Anjith George}

has received his Ph.D. and M-Tech degree from the Department of Electrical Engineering, Indian Institute of Technology (IIT) Kharagpur, India in 2012 and 2018 respectively. After Ph.D, he worked in Samsung Research Institute as a machine learning researcher. Currently, he is a post-doctoral researcher in the biometric security and privacy group at Idiap Research Institute, focusing on developing face presentation attack detection algorithms. His research interests are real-time signal and image processing, embedded systems, computer vision, machine learning with a special focus on Biometrics.
\end{IEEEbiography}

\begin{IEEEbiography}
  [{\includegraphics[width=1in,height=1.25in,clip,keepaspectratio]{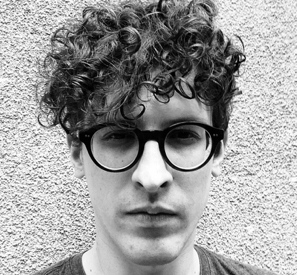}}]{David Geissb\"uhler}

 is a researcher at the Biometrics Security and Privacy (BSP) group at the Idiap Research Institute (CH) and conducts research on multispectral sensors. He obtained a Ph.D. in High-Energy Theoretical Physics from the University of Bern (Switzerland) for his research on String Theories, Duality and AdS-CFT correspondence. After his thesis, he joined consecutively the ACCES and Powder Technology (LTP) groups at EPFL, working on Material Science, Numerical Modeling and Scientific Visualisation.
\end{IEEEbiography}

\begin{IEEEbiography}
 [{\includegraphics[width=1in,height=1.25in,clip,keepaspectratio]{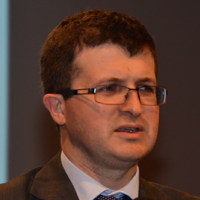}}]{S{\'e}bastien Marcel}
received the Ph.D. degree in signal processing from Universit\'{e} de Rennes I, Rennes, France, in 2000 at CNET, the Research Center of France Telecom (now Orange Labs). He is currently interested in pattern recognition and machine learning with a focus on biometrics security. He is a Senior Researcher at the Idiap Research Institute (CH), where he heads a research team and conducts research on face recognition, speaker recognition, vein recognition, and presentation attack detection (anti-spoofing). He is a Lecturer at the Ecole Polytechnique F\'{e}d\'{e}rale de Lausanne (EPFL) where he teaches a course on ``Fundamentals in Statistical Pattern Recognition.'' He is an Associate Editor of IEEE Signal Processing Letters. He has also served as Associate Editor of IEEE Transactions on Information Forensics and Security, co-editor of the ``Handbook of Biometric Anti-Spoofing,'' Guest Editor of the IEEE Transactions on Information Forensics and Security Special Issue on ``Biometric Spoofing and Countermeasures,'' and co-editor of the IEEE Signal Processing Magazine Special Issue on ``Biometric Security and Privacy.'' He was the Principal Investigator of international research projects including MOBIO (EU FP7 Mobile Biometry), TABULA RASA (EU FP7 Trusted Biometrics under Spoofing Attacks), and BEAT (EU FP7 Biometrics Evaluation and Testing).
\end{IEEEbiography}

\end{document}